\documentclass[10pt,journal,compsoc]{IEEEtran}
\usepackage[switch]{lineno}
\ifCLASSOPTIONcompsoc
  \usepackage[nocompress]{cite}
\else
  \usepackage{cite}
\fi
\ifCLASSINFOpdf
\else
\fi
\usepackage{times}
\usepackage{epsfig}
\usepackage{graphicx}
\usepackage{amsmath}
\usepackage{amssymb}

\usepackage{xcolor}

\usepackage{graphicx}
\usepackage{comment}
\usepackage{amsmath,amssymb} % define this before the line numbering.
\usepackage{color, colortbl}

\usepackage{xcolor}
\usepackage{nicefrac}
\usepackage{booktabs}
\usepackage{placeins}
\usepackage{pifont}
\usepackage{subcaption}
\usepackage{xspace}
\usepackage{arydshln}
\usepackage{nicefrac}
\usepackage{pifont}
\usepackage{multirow}

\usepackage{algorithm}
\usepackage{algpseudocode}
\algrenewcommand\algorithmicindent{1.0em}%

\usepackage{array}
\usepackage{booktabs}
\usepackage{adjustbox}

\usepackage{listings}

\usepackage[colorlinks,citecolor=red,urlcolor=blue,bookmarks=false,hypertexnames=true,pagebackref=true,breaklinks=true]{hyperref}

\definecolor{BrickRed}{HTML}{B6321C}
\definecolor{RoyalBlue}{HTML}{0071BC}
\definecolor{PineGreen}{HTML}{008B72}
\definecolor{bluefig}{HTML}{5B9BD5}
\definecolor{Gray}{gray}{0.9}

\let\thetaold\theta
\renewcommand{\theta}{\boldsymbol{\thetaold}}
\newcommand{\mcL}{\mathcal{L}}
\newcommand{\mcI}{\mathcal{I}}
\newcommand{\mcS}{\mathcal{S}}
\newcommand{\mcD}{\mathcal{D}}
\newcommand{\mcC}{\mathcal{C}}
\newcommand{\mcN}{\mathcal{N}}

\newcommand{\vx}{\mathbf{x}}
\newcommand{\vh}{\mathbf{h}}
\newcommand{\vy}{\mathbf{y}}

\DeclareMathOperator*{\argmax}{argmax} % thin space, limits underneath in displays
\newcommand{\pp}{\textit{p.p}\,}

\newcommand{\tableindent}{\,\,\,}
\newcommand{\ours}{PLOP\,}
\newcommand{\ourslong}{PLOPLong\,}

\newcommand{\cmark}{\ding{51}}%
\newcommand{\xmark}{\ding{55}}%

\begin{document}
\title{Tackling Catastrophic Forgetting and Background Shift in Continual Semantic Segmentation}

\author{Arthur Douillard\textsuperscript{1,2},
        Yifu Chen\textsuperscript{1},
        Arnaud Dapogny\textsuperscript{3},
        and~Matthieu Cord\textsuperscript{1,4}
\thanks{\textsuperscript{1} Sorbonne Université, \textsuperscript{2} Heuritech, \textsuperscript{3} Datakalab, \textsuperscript{4} Valeo.ai}
\thanks{Corresponding author: Arthur Douillard arthur.douillard@heuritech.com}
}

\IEEEtitleabstractindextext{%
\begin{abstract}
   Deep learning approaches are nowadays ubiquitously used to tackle computer vision tasks such as semantic segmentation, requiring large datasets and substantial computational power. Continual learning for semantic segmentation (CSS) is an emerging trend that consists in updating an old model by sequentially adding new classes. However, continual learning methods are usually prone to catastrophic forgetting.
   This issue is further aggravated in CSS where, at each step, old classes from previous iterations are collapsed into the background. In this paper, we propose Local POD, a multi-scale pooling distillation scheme that preserves long- and short-range spatial relationships at feature level. Furthermore, we design an entropy-based pseudo-labelling of the background w.r.t. classes predicted by the old model to deal with background shift and avoid catastrophic forgetting of the old classes. Finally, we introduce a novel rehearsal method that is particularly suited for segmentation. Our approach, called PLOP, significantly outperforms state-of-the-art methods in existing CSS scenarios, as well as in newly proposed challenging benchmarks.%\footnote{Code is available at \\\url{https://github.com/arthurdouillard/CVPR2021_PLOP}}
\end{abstract}

}

\maketitle

\IEEEdisplaynontitleabstractindextext

% \fi
\IEEEpeerreviewmaketitle

\IEEEraisesectionheading{\section{Introduction}\label{sec:introduction}}

\IEEEPARstart{C}{omputer} vision went this last decade under a dramatic change from hand-crafted features \cite{lowe1999sift,perronnin2007fisherkernels} and machine learning models \cite{cortes1995svm} to deep convolutional neural networks (DCNNS) \cite{krizhevsky2012alexnet}. Among the existing tasks involving computer vision, semantic segmentation aims to assign a label to each pixel of an image. It allows the prediction of multiple objects in the same image, and moreover their exact position and shape. This task recently florished \cite{tao2020HRNet,zhang2020resnest,chen2018ZPSA} with larger datasets with thousand of fully annotated images \cite{zhou2017adedataset,neuhold2017mapillary}, increased computational power, and larger attention \cite{wang2020axialdeeplab}. Unfortunatly, the recent research in this area is often impracticable for real-life applications: they mostly need fully annotated data and require to be retrained from scratch if a new class is added to the dataset. Idealy, one would wish to regularly expand a dataset, only adding and labelling new classes and updating the model in accordance. This setup, referred here as Continual Semantic Segmentation (CSS), has emerged very recently for specialized applications \cite{ozdemir2018learnthenewkeeptheold,ozdemir2019segmentationanotomical,tasar19incrementsegmentationremotesensing} before being proposed for general segmentation datasets \cite{michieli2019ilt,cermelli2020modelingthebackground,douillard2020plop}.

In particular, in this paper, we argue that two problems arise when performing CSS with DCNNs. The first one, inherited from continual learning, is called \textit{catastrophic forgetting}~\cite{robins1995catastrophicforgetting,french1999catastrophicforgetting,thrun1998lifelonglearning}, and points to the fact that neural networks tend to completely and abruptly forget previously learned knowledge when learning new information ~\cite{kemker2018measuringforgetting}. Catastrophic forgetting presents a real challenge for continual learning applications based on deep learning methods. For some applications, storing previously seen data is not allowed for privacy reasons. This is a real challenge because the model can no longer be trained on the old categories with labeled data. The worse part is that the previously learned categories may not even exist in the future data, which requires the model to never forget these classes in the case of learning only once. In the case where storage of past data is allowed, we can use the past data to prevent forgetting. However, given the size of the dataset, fine-tuning over all past data is extremely costly in memory and time. Therefore, rehearsal methods are usually employed, which consist of storing some of the past examples and replaying them while learning new information. This problem is exacerbated in continual segmentation where the storage cost is high and images are partially labeled.

The second challenge, specific to CSS, is the semantic shift of the background class. In a traditional semantic segmentation setup, all object categories are predefined, and the "background" class contains pixels that do not belong to any of these classes. However, in CSS, the background contains pixels that do nott belong to any of the \textit{current} classes. Thus, for a specific learning step, the background can contain both future classes, not yet seen by the model, as well as old classes. Thus, if nothing is done to distinguish pixels belonging to the real background class from old class pixels, this background shift phenomenon risks exacerbating the catastrophic forgetting even further \cite{cermelli2020modelingthebackground}. This issue also has an impact on the selection of the old data we want to store. Because some currently learned classes are annotated as background in the old data, this may degrade the performance of these classes if one naively treat them as background to fine-tune the current model.

We tackle the first challenge of catastrophic forgetting by designing a constraint enforcing a similar behavior between the old and current model. Specifically, we leverage intermediary representations of the convolutional networks to ensure that similar patterns are extracted through time. This feature-based constraints, called Local POD, fully exploits the global and local scale necessary to semantic segmentation through a multi-scale design. The second challenge, background shift, is greatly allievated a confidence-based pseudo-labeling strategy to retrieve old class pixels within the background. For instance, if a current ground truth mask only distinguishes pixels from class \texttt{sofa} and background, our approach allows to assign old classes to background pixels, e.g. classes \texttt{person}, \texttt{dog} or \texttt{background} (the semantic class). We name PLOP the model exploiting those two contributions. We then propose an extention called PLOPLong that aims to excel on long continual learning scenarios. This extention exploits cosine normalization to adapt the classifier and the Local POD resulting in an improving robustness to the discrepancy between old and new classes. Moreover, PLOPLong features a modified batch normalization which reduces the sensitivity of the model to moving statistics seen across tasks in continual learning. Finally, we investigate for the first time rehearsal learning for CSS. We propose an initial version based on rehearsing complete images which unfortunatly is memory intensive. We then create the Object Rehearsal, a rehearsal method as performant but significantly more memory efficient.

From a practical point of view, our proposed methods (PLOP, PLOPLong, and Object Rehearsal) showed three important results. First, we achieve the state-of-the-art performance on several challenging datasets. Secondly, we propose several novel scenarios to further quantify the performances of CSS methods when it comes to long term learning, class presentation order and domain shift. Last but not least, we show that our model contributions largely outperform every CSS approach in these scenarios.

To sum it up, our contributions are four-folds:
 \begin{itemize}
     \item We propose a multi-scale spatial distillation loss to better retain knowledge through the continual learning steps, by preserving long- and short-range spatial statistics, avoiding catastrophic forgetting.
     \item We introduce a confidence-based pseudo-labeling strategy to identify old classes for the current background pixels and deal with background shift.
     \item We propose PLOPLong, a carefully designed refinement of our method for dealing with long CSS scenarios. The extention comes from an adaptation of PLOP's classifier and Local POD distillation as well as batch re-normalization for better handling of both catastrophic forgetting and background shift, respectively.
     \item We design a novel memory-efficient Object rehearsal learning procedure that consists in storing and carefully pasting objects through selective erasing of foreground objects.
     \end{itemize}

Additionaly, We show that \ours significantly outperforms state-of-the-art approaches in existing scenarios and datasets for CSS, as well as in several newly proposed challenging benchmarks. Furthermore, we show that PLOPLong leads to superior performances on longer CSS scenarios. Last but not least, we show that performance of CSS models can be greatly improved where rehearsal learning is an option. In such cases, the proposed Object rehearsal allows to reach high accuracies with low memory footprint.

In this paper, we expand on our previous work published in conference venue \cite{douillard2020plop}. The updated PLOP includes both model and experimental improvements. While keeping the losses introduced with PLOP, we incorporate PLOP with the batch renormalization technique \cite{ioffe2017batchrenorm} and a cosine classifier \cite{luo2018cosine_classifier}. We structurally modify part of the network with these methods, which further mitigates forgetting on long stream of tasks compared to the original PLOP. We also investigate rehearsal learning for continual segmentation. Taking into account the background shift problem for replying and also to minimise the external memory needed to store past data, we propose a novel memory-efficient rehearsal method based on objects pasting. We extend our experiments with a class-incremental version of Cityscapes, which proved to be challenging, and we show how our model modifications handle better this dataset. 
Finally, we propose more ablations and compare rehearsal-based methods against a semi-supervised method using an external unlabeled dataset.

\section{Related Work}

Continual Semantic Segmentation is a relativery young field that started getting traction following \cite{michieli2019ilt,cermelli2020modelingthebackground}. However, this field is at the intersection of many popular topics. Therefore, we start this section with an overview of recent advances in segmentation, continual learning, and a particular aspect of the latter: rehearsal learning. We then follow with a more in-depth discussion of existing approaches to CSS.

\noindent\textbf{Semantic Segmentation} methods based on Fully Convolutional Networks (FCN)  \cite{long2015fcn,sermanet2014overfeat} have achieved impressive results on several segmentation benchmarks ~\cite{everingham2015pascalvoc, cordts2016cityscapes,zhou2017adedataset,caesar2018cocoostuff}. These methods improve the segmentation accuracy by incorporating more spatial information or exploiting contextual information specifically. Atrous convolution~\cite{chen2018deeplab,mehta2018espnet} and encoder-decoder architecture~\cite{ronneberger2015UNet,noh2015deconvolution,badrinarayanan2017segnet} are the most common methods for retaining spatial information. Examples of recent works exploiting contextual information include attention mechanisms~\cite{yuan2018ocnet,zhao2018psanet,fu2019DANet,huang2019CCNet,yuan2020ocr,tao2020HRNet,zhang2020resnest}, and fixed-scale aggregation ~\cite{zhao2017PSPNet,chen2018deeplab,chen2018ZPSA,zhang2018ContextEncoding}.

\noindent\textbf{Continual Learning} models suffers from catastrophic forgetting where old classes are forgotten, leading to a major drop of performance \cite{robins1995catastrophicforgetting,thrun1998lifelonglearning,french1999catastrophicforgetting}. Multiple approaches exist to reduce this forgetting, among them some adapt the architecture continually to integrate new knowledge \cite{yoon2018dynamically_expandable_networks,li2019learning_to_grow,yan2021der} or train together co-existing sub-networks \cite{frankle2019lottery_ticket} each specialized in one specific task \cite{fernando2017path_net,golkar2019neural_pruning, hung2019cpg}. It can be also possible to fix the knowledge drift in the classifier by recalibration \cite{wu2019bias_correction,zhao2020weightalignement,belouadah2019il2m,belouadah2020scail}.
A popular approach consists in designing constraints/distillations that enforce the new model to exhibit a similar behavior as the previous model. These constraints can be defined directly on the weights \cite{kirkpatrick2017ewc,aljundi2018MemoryAwareSynapses,chaudhry2018riemannien_walk,zenke2017synaptic_intelligence}, the gradients \cite{lopezpaz2017gem,chaudhry2019AGEM}, the output probabilities \cite{li2018lwf,rebuffi2017icarl,castro2018end_to_end_inc_learn,cermelli2020modelingthebackground}, or even intermediary representations \cite{hou2019ucir,dhar2019learning_without_memorizing_gradcam,peng2019m2kd,douillard2020podnet}.

\noindent\textbf{Rehearsal Learning} \cite{robins1995catastrophicforgetting} allows to replay a limited amount of samples from previous tasks in order to reduce forgetting. While popular for continual image classification \cite{rebuffi2017icarl,castro2018end_to_end_inc_learn,chaudhry2019tinyepisodicmemories,hayes2020remind,iscen2020incrementalfeatureadaptation,kemker2018fearnet,shin2017deep_generative_replay,liu2020mnemonics,chaudhry2019tinyepisodicmemories} or continual reinforcement learning \cite{traore2019discorl}, this approach has been fewly explored in continual segmentation. It can be explained in part because of the prohibitive memory cost of the large images used in segmentation (up to $1024 \times 2048$ vs $224 \times 224$). Another drawback comes the nature of continual segmentation where images at each step are partially labeled. This partial labelling will be different for each rehearsed image, leading to a damaging concept shift \cite{morenotorresa2012datasetshift,lesort2021driftanalysis}. Huang et al. \cite{huang2021halfrealhalffake} proposed after our PLOP a pseudo-rehearsal method for segmentation that while effective, incurs an important time overhead (up to $3\times$ longer). 

\noindent\textbf{Continual Semantic segmentation}: Despite enormous progress in the two aforementioned areas respectively, segmentation algorithms are mostly used in an offline setting, while continual learning methods generally focus on image classification. Recent works extend existing continual learning methods \cite{li2018lwf,hou2019ucir} for specialized applications \cite{ozdemir2018learnthenewkeeptheold,ozdemir2019segmentationanotomical,tasar19incrementsegmentationremotesensing} and general 
semantic segmentation \cite{michieli2019ilt}.
The latter considers that the previously learned categories are properly annotated in the images of the new dataset. This is an unrealistic assumption that fails to consider the background shift: pixels labeled as background at the current step are semantically ambiguous, in that they can contain pixels from old classes (including the real semantic background class, which is generally deciphered first) as well as pixels from future classes. Cermelli et al.~\cite{cermelli2020modelingthebackground} propose a novel classification and distillation losses. Both handle the background shift by summing respectively the old logits with the background logits and the new logits with the background. We argue that a distillation loss applied to the model output is not strong enough for catastrophic forgetting in CSS. Furthermore, their classification loss doesn't preserve enough discriminative power w.r.t the old classes when learning new classes under background shift. We introduce our PLOP framework that solves more effectively those two aspects. Yu et al.~\cite{yu2020continualsegmentationselftraining} proposed to exploit an external unlabeled dataset in order to self-training through a pseudo-labeling loss; we show that our model while not designed with this assumption in mind can outperform their performance. Cermelli et al.~\cite{cermelli2020fewshotcontinualsegm} creates a novel setting of continual \textit{few-shots} segmentation, we implement their method in our setting and draw inspiration from it to further improve PLOP. Michielli et al.\cite{michieli2021sdr} draws inspiration from the metric learning litterature to conceive a model for continual segmentation that exploits prototypes updated with an exponential moving average of the mean batch features.

\noindent\textbf{Positioning:\,} Contrary to previous works in continual segmentation \cite{michieli2019ilt,cermelli2020modelingthebackground} which reduced slightly forgetting through distillation of the probabilities, we propose a stronger constraint based on global and local statistics extracted from intermediary features. Moreover, background is often not considered \cite{michieli2019ilt} or only weakly tackled \cite{cermelli2020modelingthebackground}, while we propose to eliminate it through segmentation maps completion with pseudo-labeling. Finally, none of the work proposed rehearsal methods for CSS, while we propose a non-trivial based on image rehearsal and further improve it with a more data-efficient method based with object rehearsal. 

\section{Model}

\begin{figure*}[ht!]
\centering
    \includegraphics[width=0.8\linewidth]{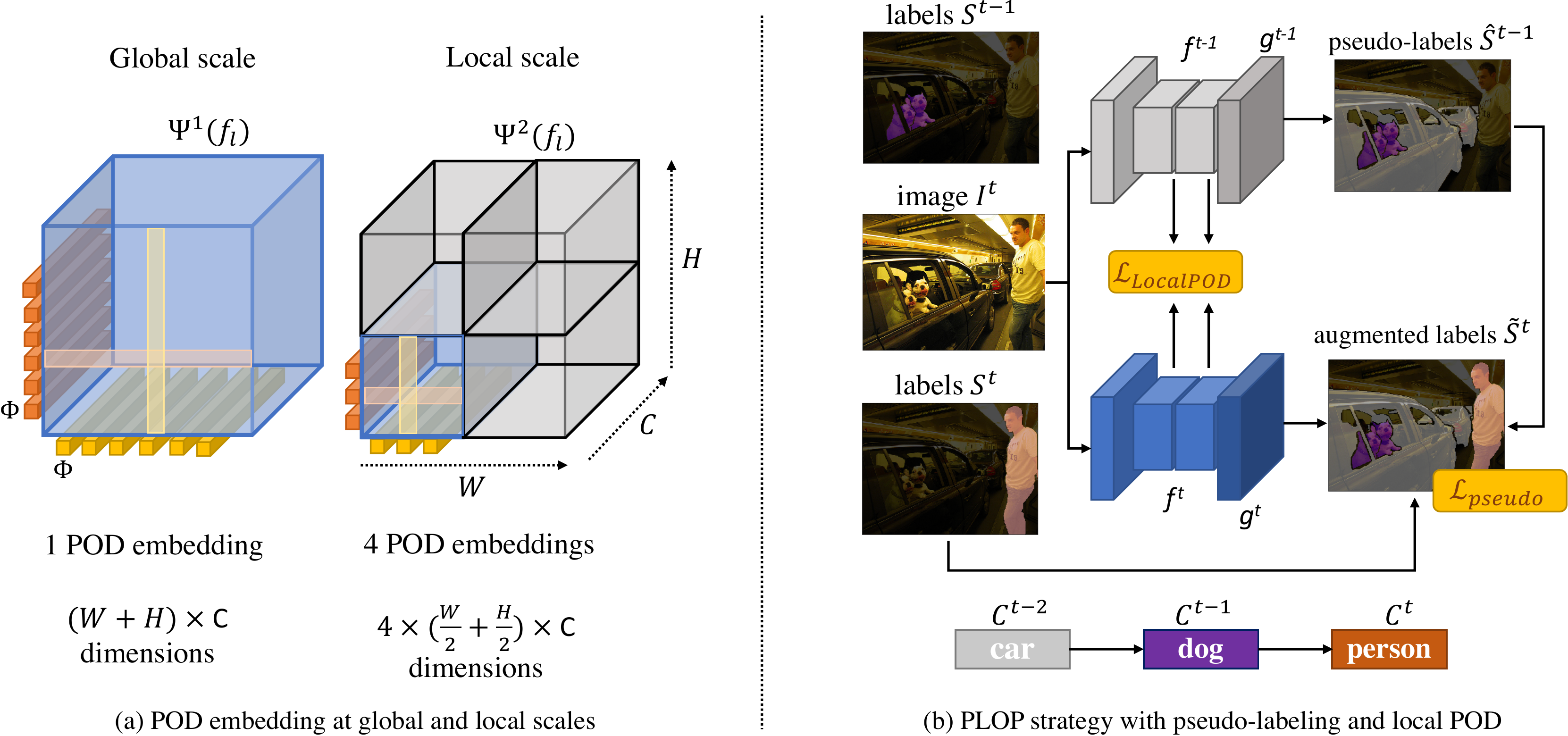} 
    \caption{Local POD details and the complete PLOP strategy. (a) Local POD consists in POD embeddings compute at multiple scale. The global scale aggregates statistics across the whole features maps while the local scale focuses on finer details.  (b) The model incrementally learns new classes (\textit{e.g.} \texttt{car}, \texttt{dog}, \texttt{person}). Only the current class (\texttt{person}) is labeled while previous classes are folded into the background. We use the previous model $g^{t-1} \circ f^{t-1}$ to generate pseudo-labels $\hat{\mcS}^{t-1}$ regarding the old classes to alleviate this ambiguity, and complete the labels $\mcS^t$ which are then used as ground-truth in $\mcL_\text{pseudo}$. The Local POD distillation is applied at multiple levels of the features extractors $f^{t-1}$ and $f^t$.} 
\label{fig:model_plop}
\end{figure*}

The model description is organized as follows: we first details the continual protocol and the notations. Then, we tackle the issue of catastrophic forgetting by designing an adapted distillation loss and we allievate the background shift by proposing a uncertainty-based pseudo-labeling. Drawing ideas from the continual learning litterature, we propose an extention of PLOP specialized for long-range continual training that we nickname PLOPLong. Finally, we detail the limits of rehearsal learning in segmentation, propose a naive adaptation to the problem, and then deliver our carefully designed method.

\subsection{Framework and notations}\label{sec:overview}

\begin{table}[t]
\centering
\caption{Notations used in this paper.}
\vspace*{-0.3cm}
\label{tab:notation}
\begin{tabular}{@{}l|c@{}}
\toprule
Total number of tasks & $T$\\
Current task & $t$\\
Current classes & $\mcC^t$ \\
Previous classes & $\mcC^{1:t-1}$\\
Future classes & $\mcC^{t+1:T}$\\
Cardinality of a set of classes & $\mcN^t=\operatorname{card}(\mcC^{t})-1$\\
Image and Segmentation maps at task $t$ & $\mcI^t$ and $\mcS^t$\\
Feature extractor at task $t$ & $f^t(\cdot)$\\
Classifier at task $t$ & $g^t(\cdot)$\\
Learnable parameters at task $t$ & $\Phi^t$\\
Predicted segmentation maps & $\hat{\mcS}^t$\\
Intermediary features at level $l$ & $f^t_l,\,l \in \{1, \dots, L\}$\\
\bottomrule
\end{tabular}
\end{table}

In Continual Semantic Segmentation (CSS), a model learns a dataset  $\mcD$ in $t=1 \dots T$ steps. At each step $t$, the model has to learn a new set of classes $\mcC^t$, for which only a subset of data $\mcD_t$, that consists in a set of pairs $(\mcI^t, \mcS^t)$, is available. $\mcI^t$ denotes an RGB input image of size $W \times H \times 3$ and $\mcS^t$ the corresponding one-hot encoded ground-truth segmentation maps of size $W \times H \times \mcC^t$. If a pixel of $\mcI^t$ belongs to either one of the previous $\mcC^{1:t-1}$ or future classes $\mcC^{t+1:T}$, it is not labeled as such in $\mcS^t$ and shall instead be considered as belonging to the background class, denoted $c_\text{bg}$. However, at test time, a model at step $t$ must be able to discriminate between all the classes that have been seen so far, \textit{i.e.} $\mcC^{1:t}$. This leads us to identify two major pitafalls in CSS: the first one, inherited from continual learning, is catastrophic forgetting \cite{robins1995catastrophicforgetting,french1999catastrophicforgetting}. It suggests that a network will completely forget the old classes $\mathcal{C}^{1:t-1}$ when learning the new ones $\mathcal{C}^t$. Furthermore, catastrophic forgetting is aggravated by the second pitfall, specific to CSS, that we call background shift: at step $t$, the pixels labeled as background are indeed ambiguous, as they may contain either old (including the real background class, predicted in $\mathcal{C}^{1}$) or future classes.

We define our model at step $t$ as a composition of a feature extractor $f^t(\cdot)$ (a ResNet 101 \cite{he2016resnet} backbone and a classifier $g^t(\cdot)$. The output predicted segmentation map can then be written $\hat{\mcS}^t = g^t \circ f^t(\mcI)$. We denote the intermediate features at each layer of the feature extractor $f_l^t(\cdot)\,,\, l \in \{1, \dots L\}$. Finally, we denote the set of learnable parameters of $f^t$ and $g^t$ as $\Theta^t$. The notations introduced in this paper are summarized in \autoref{tab:notation}.

\subsection{Overcoming catastrophic forgetting in CSS with local distillation}\label{sec:distillation}

In this section, we propose to tackle the issue of catastrophic forgetting in continual learning in general and in CSS in particular. An effective method for doing so involves setting constraints between the old ($g^{t-1} \circ f^{t-1}$) and current ($g^{t} \circ f^{t}$) models. These constraints aim at enforcing a similar behavior between both models and in turn reduce the loss of performance on old classes. A common such contraint is based on applying knowledge distillation \cite{hinton2015knowledge_distillation,li2018lwf} between the predicted probabilities of both models. When applied to CSS, such distillation loss must be carefully balanced to find a good trade-off between rigidity (\textit{i.e.} too strong constraints, resulting in not being able to learn new classes) and plasticity (\textit{i.e.} enforcing loose constraints, which can lead to catastrophic forgetting of the old classes).

In previous work \cite{douillard2020podnet} we designed POD, short for Pooled Output Distillation. Rather than solely constraining the output model probabilities, POD enforces consistenty between intermediary statistics of both models. In practice, for a feature map $\vx$, we define a POD embedding $\Phi(\cdot)$ as:

\begin{equation}
    \Phi(\vx) = \left[\frac{1}{W} \sum_{w=1}^W \vx[:,w,:] \bigg\Vert \frac{1}{H} \sum_{h=1}^H \vx[h,:,:]\right] \in \mathbb{R}^{(H + W) \times C}\,,
\label{eq:pod_embedding}
\end{equation}

where $[\cdot\,\|\,\cdot]$ denotes concatenation over the channel axis. The POD embedding is thus computed as the concatenation of the $H \times C$ width-pooled slices and the $W \times C$ height-pooled slices of $\vx$ and captures long-range statistics across the whole features maps. The POD distillation loss then consists in minimizing the $\mathcal{L}_2$ distance between POD embeddings computed at several layers $l \in \{1, \dots, L\}$, w.r.t the current model parameters $\Theta^t$:

\begin{equation}
    \mcL_\text{pod}(\Theta^t) = \frac{1}{L} \sum_{l = 1}^L \left\Vert  \Phi(f^t_l(\mcI)) -  \Phi(f^{t-1}_l(\mcI)) \right\Vert^2\,.
\label{eq:pod_loss}
\end{equation}

POD yields state-of-the-art results in continual learning for image classification, especially when large numbers of tasks are considered, a case where the aforementioned plasticity-rigidity trade-off becomes even more crucial. Another interest arises in the context of CSS: the long-range statistics computed across an entire axis (horizontal or vertical) which reminds recent work on attention for segmentation \cite{wang2020axialdeeplab,huang2020ccnet,park2020csc} which aim to enlarge the receptive field through global attention/statistics \cite{wang2020axialdeeplab}.

In the frame of classification, it is, to a certain extent, necessary to discard spatial information through global pooling. However, conversely, semantic segmentation requires preservation of both long-range and short-range statistics, making a distillation loss such as POD suboptimal for that purpose.

Following this reflexion, and inspired by the multi-scale litterature \cite{lazbnik2006spatial_pyramid_matching,he2014spatialpyramidpooling}, we design a distillation loss, called Local POD that retain the long-range spatial statistics while also preserving the local information. The proposed Local POD consists in computing the width and height statistics at different scales $\{1/2^s\}_{s=0 \dots S}$, as illustrated in \autoref{fig:model_plop} (a). At a given level $l$ of the features extractor, $s^2$ POD embeddings are computed per scale $s$ and concatenated:

\begin{equation}
    \Psi^s(\vx) = \left[ \Phi(\vx^s_{0,0}) \| \dots \| \Phi(\vx^s_{s-1,s-1}) \right] \in \mathbb{R}^{S \times (H + W) \times C}\,,
\label{eq:localpod_embedding1}
\end{equation}

where $\forall i = 0 \dots s-1$, $\forall j = 0 \dots s-1$, $\vx^s_{i,j} = \vx[i H/s:(i+1) H/s, j W/s:(j+1) W/s,:]$ is a sub-region of the embedding tensor $\vx$ of size $W/s \times H/s$. We then concatenate (along the channel axis) the Local POD embeddings $\Psi^s(\vx)$ of each scale $s$ to form the final embedding:

\begin{equation}
    \Psi(\vx) = \left[ \Psi^1(\vx) \| \dots \| \Psi^S(\vx) \right] \in \mathbb{R}^{(1 + \dots + S) \times (H + W) \times C}\,.
\label{eq:local_pod}
\end{equation}

Similarly to POD, we compute Local POD embeddings for every layer $l \in \{1, \dots, L\}$ of both the old and current models. The resulting loss is thus:

\begin{equation}
    \mcL_{\scriptstyle\text{LocalPod}}(\Theta^t) = \frac{1}{L} \sum_{l = 1}^L \left\Vert  {\Psi}(f^t_l(I)) -  {\Psi}(f^{t-1}_l(I)) \right\Vert^2\,.
\label{eq:local_pod_loss}
\end{equation}

Thus, notice that the first scale of Local POD ($1/2^0$) is similar to the original POD and models long-range dependencies across the entire image. The subsequent scales ($s=1/2^1, 1/2^2 \dots$), enforce short-range dependencies. Thus, the proposed Local POD tackles the problem of catastrophic forgetting by modeling and preserving long and short-range statistics between the old and current models, throughout the CSS steps. 

\subsection{Pseudo-labeling to fix background shift}\label{sec:hardpl}

In addition to catastrophic forgetting, a successful CSS approach shall handle the background shift problem, thus shall take into account the ambiguity of pixels labelled as background at each step. We propose a pseudo-labeling strategy that ``\textit{completes}'' the ambiguous background labels. Pseudo-labeling \cite{lee2013pseudolabel} is commonly used in domain adaption for semantic segmentation \cite{vu2019advent,li2019bidirectionallearning,zou2018classbalancedselftraining,saporta2020esl} where a model is trained to match both the labels of a source dataset and the pseudo-labels (usually obtained using the same predictive model, in a self-training fashion) of an unlabeled target dataset. In this case, the knowledge acquired on the source dataset helps the model to generate labels for the target dataset. In the frame of CSS, at each step, we use the predictions of the old model to decipher previously seen classes among the ambiguous background pixels, as illustrated in \autoref{fig:model_plop}. The pseudo-labeling rely on the previous model which can be uncertain for some pixels due to inherent bias to the optimization and because of the forgetting. Therefore in order to avoid propagate errors through incorrect pseudo-labels, we filter out the most uncertain ones based on a adaptive entropy-based threshold.

Formally, let $\mcN^t=\operatorname{card}(\mcC^{t})-1$ the cardinality of the current classes excluding the background class. Let $\hat{\mcS}^{t} \in \mathbb{R}^{W,H,1+\mcN^1 +\dots + \mcN^t}$ denotes the predictions of the current model (which includes the real background class, all the old classes as well as the current ones). We define $\tilde{\mcS}^{t} \in \mathbb{R}^{W,H,1+\mcN^1 +\dots + \mcN^t}$ the target as step $t$, computed using the one-hot ground-truth segmentation map $\mcS^{t} \in \mathbb{N}^{W,H,1+\mcN^t}$ at step $t$ as well as pseudo-labels extracted using the old model predictions $\hat{\mcS}^{t-1} \in \mathbb{R}^{W,H,1+\mcN^1 +\dots + \mcN^{t-1}}$ as follows:

\begin{equation}
    \footnotesize
    \tilde \mcS^{t}\left(w,h,c\right)= \mkern-5mu \left\{\begin{array}{ll}
    \mkern-10mu 1 \mkern-27mu & \text { if } S^{t} (w,h,c_{bg})=0 \text { and } c = \argmax \limits_{c' \in \mcC^{t}} \mcS^{t}(w,h,c') \\
    \mkern-10mu 1 \mkern-27mu & \text { if } \mcS^{t}(w,h,c_{bg})=1 \text { and } c = \mkern-8mu \argmax \limits_{c' \in \mcC^{1:t-1}} \mkern-6mu \hat{\mcS}^{t-1}(w,h,c') \\
    \mkern-10mu 0 \mkern-27mu & \text { otherwise }\\
    \end{array}\right.
    \label{eq:pseudo_bis}
\end{equation}

In other words, in the case of non-background pixels we simply copy the ground truth label. Otherwise, we use the class predicted by the old model $g^{t-1}(f^{t-1}(\cdot))$. This pseudo-label strategy allows to assign each pixel labelled as background his real semantic label if this pixel belongs to any of the old classes. However, pseudo-labeling all background pixels can be unproductive, e.g. on uncertain pixels where the old model is likely to fail. Therefore we only keep pseudo-labels where the old model is deemed ``\textit{confident}'' enough. \autoref{eq:pseudo_bis} is modified to take into account this uncertainty:

\begin{equation}
    \scriptsize
    \tilde \mcS^{t}\left(w,h,c\right)\mkern-5mu = \mkern-5mu \left\{\begin{array}{ll}
    \mkern-14mu 1 \mkern-27mu & \mkern-8mu \text { if } \mcS^{t} \mkern-4mu (w,h,c_{bg})\mkern-4mu = \mkern-4mu 0 \text { and } c \mkern-4mu = \mkern-4mu \argmax \limits_{c' \in \mathcal{C}^{t}} \mcS^{t} \mkern-4mu (w,h,c') \\
    \mkern-14mu 1 \mkern-27mu & \mkern-8mu \text { if } \mcS^{t} \mkern-4mu (w,h,c_{bg}) \mkern-4mu = \mkern-4mu 1 \text { and } c \mkern-4mu = \mkern-12mu \argmax \limits_{c' \in \mathcal{C}^{1:t-1}} \mkern-6mu \hat \mcS^{t-1} \mkern-4mu (w,h,c') \text{ and } u \mkern-4mu < \mkern-4mu \tau_{c}\\
    \mkern-14mu 0 \mkern-27mu & \mkern-4mu \,\text { otherwise\,, }\\
    \end{array}\right.
    \label{eq:pseudo_bis_uncertain}
\end{equation}

By notation abuse, $u$ is function $u(\mcS^t(w, h))$ that measures the uncertainty of the current model given a pixel $\mcI(w,h)$. $\tau_{c}$ denotes a class-specific uncertainty threshold. Hence, in the case where the old model is uncertain ($u \ge \tau_c$) about some pixels, they will be ignored in the final classification loss. Our framework is agnostic to the type of uncertainty used, but in practice we define it as the entropy. Therefore we use for $u$ the current model's per-pixel entropy $u(\mcS^t(w, h)) = -\sum_{c \in C^{1:t}} \mcS^t(w, h, c) \log \mcS^t(w, h, c)$. Likewise, the class-specific threshold $\tau_c$ is computed from the median entropy of the old model over all pixels of $\mcD^t$ predicted the class $c$ for all $c \in \mcC^{1:t-1}$ as proposed in \cite{saporta2020esl}. Consequently, the cross-entropy loss with pseudo-labeling of the old classes can be written as:

\begin{equation}
    \mcL_\text{pseudo}(\Theta^t)=- \frac{\nu}{WH} \sum_{w,h}^{W,H} \sum_{c \in \mathcal{C}^{t}} \tilde \mcS\left(w,h,c\right) \log \hat \mcS^{t}\left(w,h,c\right)\,.
\label{eq:pseudo_loss}
\end{equation}

We reduce the normalization factor $WH$ by as much discarded pixels. To avoid giving disproportional importance to the pixels belonging to new classes (which aren't discarded), we introduce in \autoref{eq:pseudo_loss} an daptive factor $\nu$, which is the ratio of accepted old classes pixels over the total number of such pixels. \textit{i.e.} if most if most of the image is uncertain, the overall importance of the image relative to other images in the batch is reduced. The overall behavior of our pseudo-labeling is illustrated in \autoref{fig:model_plop}.

We call our final model PLOP (standing for Pseudo-labeling and LOcal Pod). PLOP's final loss is a weighted combination of \autoref{eq:local_pod_loss} and \autoref{eq:pseudo_loss}: 

\begin{equation}
    \mcL(\Theta^t) = \underbrace{\strut \mcL_\text{pseudo}(\Theta^t)}_\text{classification} + \lambda\underbrace{\strut \mcL_\text{localPod}(\Theta^t)}_\text{distillation}\,,
\label{eq:complete_loss}
\end{equation}

with $\lambda$ an hyperparameter. PLOP, while already very competitive, can face difficulties when dealing with long continual settings, \textit{i.e.} for which the number of steps grows larger. For this reason, we propose PLOPLong, an extension of PLOP for dealing with such cases.

\subsection{PLOPLong: a specialization for long settings}\label{sec:plopv2}

A first axis of improvement consists in correcting the forgetting located at the classifier level. Indeed, the classifier weights tend to be specialized for the last classes to the detriment of older classes \cite{hou2019ucir}. Several means exist to correct this bias \cite{wu2019bias_correction,belouadah2019il2m,zhao2020weightalignement,luo2018cosine_classifier} and we choose the cosine normalization. In practice, we replace the classifier by a cosine classifier \cite{luo2018cosine_classifier}, where the final inner product is discarded in favor of a cosine similarity. By doing so, all class weights --both old and new-- have a constant magnitude of 1, which reduces drastically the bias towards new classes. The classifier $g^t$ is in segmentation a pointwise ($1\times1$ kernel) convolution which doesn't alter the spatial organisation but maps the $ch$ features channels to $\mcC^{1:t}$ channels, one per class to predict. This pointwise convolution can be seen as a fully-connected layer that is apply independentely to each pixel. Therefore, the classifier $g^t$ has $\{\theta_c^t \in \mathbb{R}^{ch} | \forall c \in \mcC^{1:t}\}$. The cosine normalization can then be express as:

\begin{equation}
\hat{\mcS}(w, h, c) = \frac{\alpha \langle \theta^t_c, \vh(w, h) \rangle}{\Vert \theta^t_c \Vert^2 \Vert \vh(w, h) \Vert^2}\,
\label{eq:cosine_classifier}
\end{equation}

with $\alpha$ a learned scalar parameter initialized to 1 and helping the convergence, and $\vh \in \mathbb{R}^{W \times H \times ch}$ the final features embedding before the classifier. First used in continual learning by Hou et al.~\cite{hou2019ucir}, this classifier has more recently be adopted by continual few-shot segmentation \cite{cermelli2020fewshotcontinualsegm} or multi-modes continual learning \cite{douillard2020podnet}. New cosine classifier weights for new classifier can then be initialized with weight imprinting \cite{qi2018imprintedweights} as recently adapted for segmentation by Cermelli et al.~\cite{cermelli2020fewshotcontinualsegm}.

As second improvement we propose to exploit the cosine normalization for intermediary features: the comparison between the Local POD embeddings of the previous model $f^{t-1}$ with the current $f^t$ can be too constrained. We relax the constraints by imposing not a low L2 distance between both embeddings, but rather a high cosine similarity. We simply need to alter \autoref{eq:local_pod_loss} by replacing the $\Phi(\cdot)$ operator by $\bar{\Phi}(\cdot) = \nicefrac{\Phi(\cdot)}{\vert\Phi(\cdot)\Vert^2}$. However note that the features from level $l$ given to level $l+1$ are not normalization. The normalization only happens for the Local POD embeddings. 

\begin{figure*}[ht!]
\centering
    \includegraphics[width=1.0\linewidth]{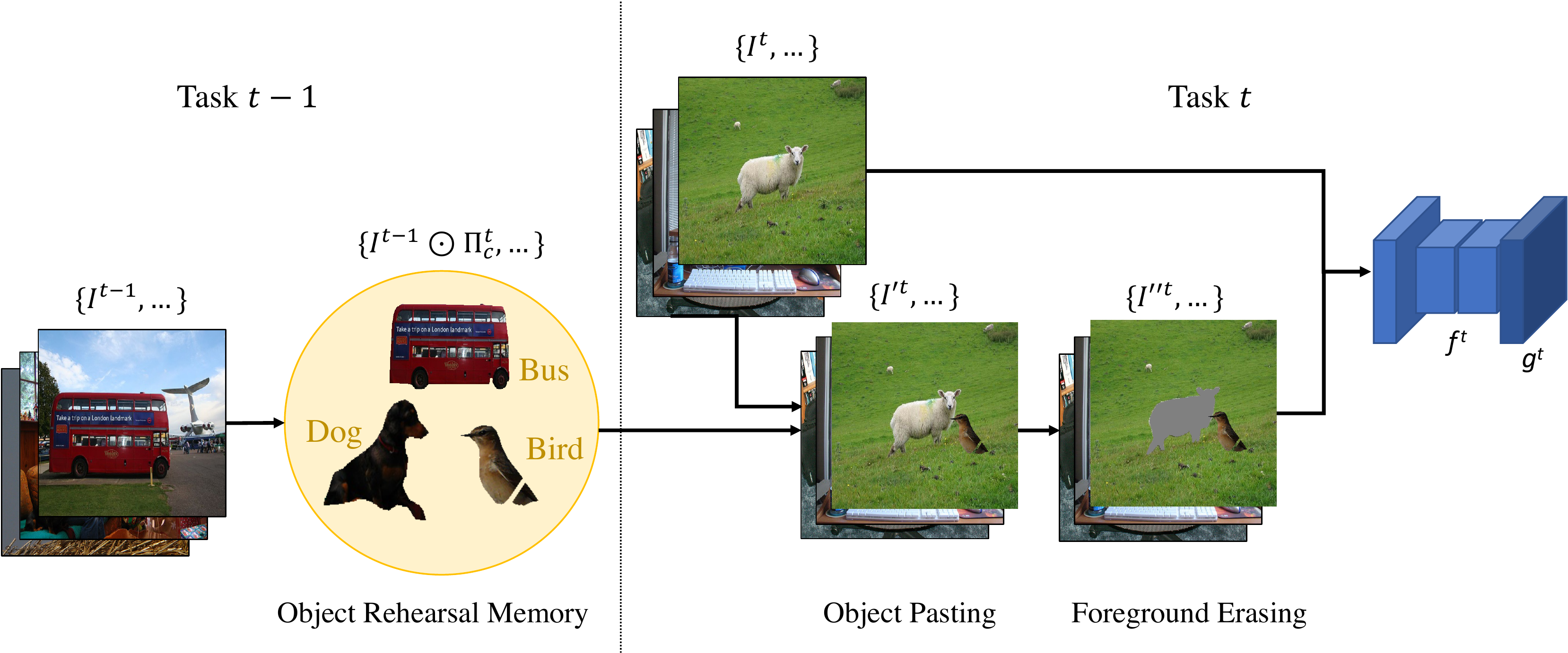} 
    \caption{Our Object Rehearsal strategy. In task $t-1$, we select from $\{\mcI^{t-1}, \dots \}$ a limited amount of objects (here \texttt{bus}, \texttt{bird}, and \texttt{dog}), which will be then mixed in the images $\{\mcI^{t}, \dots \}$ from the current task $t$; after pasting, the other present objects are erased. Finally the current model $g^t \circ f^t$ will be given the concatenation of the the original images and the augmented images $\{\mcI''^t, \dots\}$.}
\label{fig:model_objectrehearsal}
\end{figure*} 

The third axis of improvement focuses the Batch Normalization \cite{ioffe2015batchnorm}, an almost ubiquitous component to deep computer vision models. It normalizes internal representation with the batch mean $\mu_\mathbb{B}$ and batch standard deviation $\sigma_\mathbb{B}$ (resp. running mean $\mu$ and running std $\sigma$) during training (resp. testing). Formally for a all batch normalization layer taking an input $\vx$ and producing an output $\vy$:

\begin{equation}
    \vy = \frac{\vx - \mu_\mathbb{B}}{\sigma_\mathbb{B}} \cdot \gamma + \beta
\label{eq:batch_norm}
\end{equation}

with $\gamma$ and $\beta$ two learned parameters. The drawback of this normalization layeris its assumption that the data is sampled \textit{i.i.d.} which is not the case in continual segmentation where multiple shifts \cite{morenotorresa2012datasetshift,lesort2021driftanalysis,douillardlesort2021continuum}, happen in the training tasks, different from the testing tasks. Lomonaco et al.~\cite{lomonaco2020ar1} and Cermelli et al.~\cite{cermelli2020fewshotcontinualsegm} proposed to replace the batch normalization by the batch \textit{re}normalization \cite{ioffe2017batchrenorm}:

\begin{equation}
    \vy = \frac{\vx - \mu_\mathbb{B}}{\sigma_\mathbb{B}} \cdot r + d,\,\text{where}\, r = \frac{\sigma_\mathbb{B}}{\sigma},\, d = \frac{\mu_\mathbb{B} - \mu}{\sigma}\,
\label{eq:batch_renorm}
\end{equation}

Intuitively, batch renormalization avoids the discrepancy between training and testing of the batch normalization. Furthermore, following \cite{cermelli2020fewshotcontinualsegm}, we freeze during training the statistics ($\mu$ and $\sigma$) after the first task to avoid harmful statistics drifts.

All three improvements further increase performance for long series of tasks: (1) Cosine classifer reduces the increased bias between recent and old classes. (2) A cosine-based Local POD relax the constraints in order to learn correctly the bigger number of new classes. (3) Frozen BatchReNorm reduces the inherent drift of statistics that grew larger as the number of tasks increase.

\subsection{Rehearsing previous data for CSS}\label{sec:object_rehearsal}

Neither the proposed PLOP or PLOPLong did make use of previously seen data $\{\mcD^{1}, \dots, \mcD^{t-1}\}$ when considering step $t$. In this section, we explore how to further improve CSS performance if a model is now allowed to rehearse
a limited amount of previous data. We first consider a traditional approach, namely Image Rehearsal, that we adapt to the CSS setting, and then propose a more efficient method, that we call Object Rehearsal. 

\subsubsection{Image Rehearsal}

We first consider rehearsing a limited amount of images from previous tasks during the current task. Contrary to image classification, in CSS, images can have multiple labels if several objects (e.g. \texttt{car}, \texttt{sky}) are present in the image. We therefore propose to select $M$ images for each class $c \in C^t$, resulting in $M \times |C^t|$. We ensure that the selected images are unique, thus because multiple classes can co-exist in the same image, the resulting amount of sampled classes may be above $M \times |C^t|$.
We select the images with a simple random selection, which has been proved to generally be as good as more elaborate methods \cite{castro2018end_to_end_inc_learn}. The memory footprint will grow until all tasks but the last are seen, resulting in a total amount of $M \times |C^{1:T-1}|$ stored images. Note however that a naive rehearsal will only bring marginal gains because the stored ground truth segmentation maps are partially labeled. Indeed, images stored from a previous task $t_o$ ($t_o < t$, with $t$ the current task) will only have the classes $C^{t_o}$ labeled, and the other classes folded into the background. Here again, pseudo-labeling (see \autoref{sec:hardpl}) can \textit{complete} the segmentation maps with both even older classes $C^{1:t_o-1}$ and newer classes $C^{t_o+1:t}$.

\subsubsection{Object Rehearsal}

The main drawback of image rehearsal is that CSS images are usually large ($512\times 512$ to $1024 \times 2048$) yet they are sparsely informative, as a significant part of the images consists in background pixels \cite{lin2017focalloss} (e.g. 63\% of Pascal-VOC \cite{everingham2015pascalvoc} pixels) or belongs to a majoritary class (e.g. 32\% of Cityscapes \cite{cordts2016cityscapes} pixels are \texttt{roads}).

To address both problems, instead of storing whole images from the previous tasks $\{1, ..., t-1\}$, we propose to store an informative portion that we will mix with the images of the current task $t$. Image mixing is popular for classification \cite{hingyi2018mixup,yun2019cutmix,dabouei2020supermix,verma2019manifoldmixup,li2021moex,rame2021mixmo} yet, to the best of our knowledge, sees limited use for semantic segmentation \cite{fang2019instaboost,olsson2021classmix,zhang2021objectaug,tranheden2021dacs,ghiasi2020simplecopypaste}, and has never been considered to design memory-efficient rehearsal learning systems. Formally, given an image $I$ and the corresponding ground truth segmentation maps $S^t$, we define a binary mask $\Pi_c$  such that $\forall c \in C^t$:

\begin{equation}
    O_c = I \odot \Pi_c\,\text{where}\,\Pi_c(x, y) = 
    \begin{cases}
    1 & \text { if } S^t(x, y) = c \\
    0 & \text { otherwise }
\end{cases},
\label{eq:mask_object}
\end{equation}

\noindent with $O_c$ the selected object for class $c$. By nature, this patch is extremely sparse and can be efficiently stored on disk by modern compression algorithms \cite{ISO10918}. The total memory footpring at task t is thus $M \times |C^{1:t}|$ objects.

Then, when learning a task $t > 1$, the model will learn on both the task dataset $\cup_i^n (I^t_i, S^t_i)$ and the object memory $\cup_{c \in C^{1:t-1}} \cup_i^M O_{c,i}$. The latter will augment the former through object pasting. We augment each object by applying an affine transformation matrix \cite{fang2019instaboost}:

\begin{equation}
    \mathbf{T}=\left[\begin{array}{ccc}
    z \cos \alpha & z \sin \alpha & 0 \\
    -z \sin \alpha & z \cos \alpha & 0 \\
    0 & 0 & 1
    \end{array}\right]\,,
\label{eq:transformation_matrix_complex}
\end{equation}

\noindent with $z$ a zoom factor, and $\alpha$ an in-plane rotation angle. Note that we don't translate the object as its original position is often a good prior: indeed, objects are usually located at the same location (e.g. pedestrian on the left and right sidewalk, car on the middle road, etc.) We abuse the notation by denoting $T(.)$ the application of this transformation matrix. Once the data augmentation in done on the object (and its mask), we paste it on an image $I^t$ that refers to the current task:

\begin{equation}
    {I'}^t = I^t \odot (\mathbf{1} - T(\Pi_c)) + T(O_c) \odot T(\Pi_c)\,.
\label{eq:pasting}
\end{equation}

The pasting can result in local incoherence where the pixel of a \texttt{cow} is pasted on top or next to the pixel of a \texttt{television}. Naively the object borders can be smoothed into the image with a Gaussian filter. Unfortunatly it results in inprecise countours which are important in segmentation \cite{chen2020semeda}.

This approach has already been envisoned as a form of data augmentation in semantic segmentation, although the gains were small and to avoid the noise induced by this pasting, the batch size is prohibitively large (up to 512) \cite{ghiasi2020simplecopypaste}. We propose to reduce interference between the pasted object and the destination images by \textbf{selective erasing} of the surrounding pixels.
Indeed, given a binary matrix $\Xi(\Pi, S)$ of the same dimension as $I$ and $\Pi$:

\begin{equation}
    {I''}^t = {I'}^t \odot (\mathbf{1} - \Xi) + \kappa \odot \Xi\,.
\label{eq:erasing_pixel}
\end{equation}

We replace the pixels erased according to the mask $\Xi$ with a RGB color $\kappa \in \mathbb{R}^3$. This RGB vector could be chosen through in-painting \cite{fang2019instaboost} or be random noise, but in practice we choose a constant color gray. We also update accordingly the segmentation maps likewise:

\begin{equation}
    {S''}^t = {S''}^t \odot (\mathbf{1} - \Xi) + 255 \odot \Xi\,,
\label{eq:erasing_label}
\end{equation}

where \texttt{255} is the default value used in segmentation to ignore some pixel labels which won't be counted in the classification loss \autoref{eq:pseudo_loss}. We illustrate our rehearsal strategy in \autoref{fig:model_objectrehearsal}.

\begin{table}[t]
\centering
\caption{Description of the three datasets considered in this paper. For datasets without explicit background class, one is created based on unlabeled pixels.}
\vspace*{-0.3cm}
\label{tab:dataset_description}
\begin{tabular}{@{}l|ccccc@{}}
\toprule
Dataset & \# classes & {\scriptsize Background?} & \# train & \# test & {\scriptsize Image size} \\
\midrule
Pascal-VOC \cite{everingham2015pascalvoc} & 20 & \cmark & 10k & 1.5k & $512 \times 512$\\
Cityscapes \cite{cordts2016cityscapes} & 19 & \xmark & 3k & 0.5k & $512 \times 1024$\\
ADE20k \cite{zhou2017adedataset} & 150 & \xmark & 20k & 2.0k & $512 \times 512$\\
\bottomrule
\end{tabular}
\end{table}

\begin{table}[t]
\centering
\caption{Description of the 12 different benchmarks evaluated in this paper. For some of these, each task brings new classes while, for others, it comes with new domains.}
\vspace*{-0.3cm}
\label{tab:setting_description}
\begin{tabular}{@{}l|rcccc@{}}
\toprule
{\scriptsize Dataset} & {\scriptsize Setting}  & {\scriptsize Mode} & {\scriptsize \# tasks} & {\scriptsize \# base classes} & {\scriptsize \# classes / inc. task} \\
\midrule
\multirow{4}{*}{Pascal-VOC} & 19-1 & class & 2 & 19 & 1 \\
                            & 15-5 & class & 2 & 15 & 5 \\
                            & 15-1 & class & 6 & 15 & 1 \\
                            & 10-1 & class & 11 & 10 & 1 \\
\hline
\multirow{1}{*}{Cityscapes} & 14-1 & class  & 6 & 14 & 1 \\
                            %& 11-5 & domain  & 3 & 11 & 5 \\
                            %& 11-1 & domain  & 11 & 11 & 1 \\
                            %& 1-1 & domain  & 21 & 1 & 1 \\
\hline
\multirow{4}{*}{ADE20k}     & 100-50 & class & 2 & 100 & 50 \\
                            & 50-50 & class  & 2 & 50 & 50 \\
                            & 100-10 & class  & 6 & 100 & 10 \\
                            & 100-5 & class  & 11 & 100 & 5 \\
\bottomrule
\end{tabular}
\end{table}

\begin{comment}
\begin{figure}
  \centering
  \includegraphics[width=0.7\linewidth]{images/dataset_viz.pdf} 
     \vspace*{-0.3cm}
    \caption{Visualization of an example image and its segmentation maps for Pascal-VOC, ADE20k, and Cityscapes.}
    \label{fig:dataset_viz}
\end{figure}
\end{comment}

\begin{table*}[t]
\centering
\caption{Continual Semantic Segmentation results on Pascal-VOC 2012 in Mean IoU (\%). $\dagger$: results excerpted from ~\cite{cermelli2020modelingthebackground}, $\diamond$ from \cite{michieli2021sdr}. Other results comes from re-implementation.}
\vspace*{-0.3cm}
\label{tab:voc_sota1}
\begin{tabular}{@{}l|cccc||cccc||cccc@{}}
\toprule
& \multicolumn{4}{c}{\textbf{19-1} (2 tasks)} & \multicolumn{4}{c}{\textbf{15-5} (2 tasks)} & \multicolumn{4}{c}{\textbf{15-1} (6 tasks)}\\
\textbf{Method} & 0-19 & 20 & \textit{all} & \textit{avg} & 0-15 & 16-20 & \textit{all} & \textit{avg} & 0-15 & 16-20 & \textit{all} & \textit{avg}\\
\midrule
% from paper MiB
$\text{Fine Tuning}^\dagger$ & \tableindent 6.80 & 12.90 & \tableindent 7.10 &  & \tableindent 2.10 & 33.10 & \tableindent 9.80 &  & \tableindent 0.20 & \tableindent 1.80 & \tableindent 0.60 & \\
$\text{PI}^\dagger$ \cite{zenke2017synaptic_intelligence} & \tableindent 7.50 & 14.00 & \tableindent 7.80 &  & \tableindent 1.60 & 33.30 & \tableindent 9.50 &  & \tableindent 0.00 & \tableindent 1.80 & \tableindent 0.50 & \\
$\text{EWC}^\dagger$ \cite{kirkpatrick2017ewc} & 26.90 & 14.00 & 26.30 &  & 24.30 & 35.50 & 27.10 &  & \tableindent 0.30 & \tableindent 4.30 & \tableindent 1.30 &  \\
$\text{RW}^\dagger$ \cite{chaudhry2018riemannien_walk} & 23.30 & 14.20 & 22.90 &  & 16.60 & 34.90 & 21.20 &  & \tableindent 0.00 & \tableindent 5.20 & \tableindent 1.30 & \\
$\text{LwF}^\dagger$ \cite{li2018lwf} & 51.20 & \tableindent 8.50 & 49.10 &  & 58.90 & 36.60 & 53.30 &  & \tableindent 1.00 & \tableindent 3.90 & \tableindent 1.80 & \\
$\text{LwF-MC}^\dagger$ \cite{rebuffi2017icarl} & 64.40 & 13.30 & 61.90 &  & 58.10 & 35.00 & 52.30 &  & \tableindent 6.40 & \tableindent 8.40 & \tableindent 6.90 & \\
$\text{ILT}^\dagger$ \cite{michieli2019ilt} & 67.10 & 12.30 & 64.40 &  & 66.30 & 40.60 & 59.90 &  & \tableindent 4.90 & \tableindent 7.80 & \tableindent 5.70 & \\ 
$\text{ILT}$ \cite{michieli2019ilt} & 67.75 & 10.88 & 65.05 & 71.23 & 67.08 & 39.23 & 60.45 & 70.37 & \tableindent 8.75 & \tableindent 7.99 & \tableindent 8.56 & 40.16 \\ 

$\text{MiB}^\dagger$ \cite{cermelli2020modelingthebackground} & 70.20 & 22.10 & 67.80 &    & 75.50 & 49.40 & 69.00 &  & 35.10 & 13.50 & 29.70 & \\
% from us
MiB \cite{cermelli2020modelingthebackground} & 71.43 & 23.59 & 69.15  & 73.28  & \textbf{76.37}  & 49.97  & \textbf{70.08} & \textbf{75.12} & 34.22 & 13.50  & 29.29  & 54.19 \\
$\text{SDR}^\diamond$ \cite{michieli2021sdr} & 71.30 & 23.40 & 69.0 & & 76.30 & 50.20 & 70.10 & & 47.30 & 14.70 & 39.50 & \\
GIFS \cite{cermelli2020fewshotcontinualsegm} & 57.88 & 32.82 & 56.69 & 67.05 & 23.61 & 16.43 & 21.90 & 50.97 & 59.36 & 13.89 & 48.53 & 61.43 \\
\ours & \textbf{75.35} & 37.35 & \textbf{73.54} & \textbf{75.47} & 75.73 & \textbf{51.71} & \textbf{70.09} & \textbf{75.19} & 65.12 & 21.11 & 54.64 & 67.21\\
\ourslong & 74.75 & \textbf{39.68} & 73.08 & 74.32 & 75.95 & 48.31 & 69.37 & 73.58 & \textbf{72.00} & \textbf{26.66} & \textbf{61.20} & \textbf{70.02}\\
%{\color{red}HRHF} \cite{huang2021halfrealhalffake} & 76.60 & 57.30 & 75.70 & & 78.90 & 57.80 & 73.90 & & 72.40 & 39.60 & 64.60 & \\
%\midrule
% Algo &   &   &   &   &   &  &  &   &   &   &  &  \\
%\midrule
%Joint model & 77.40 & 78.00 & 77.40 & --- & 79.10 & 72.60 & 77.40 & --- & 79.10 & 72.60 & 77.40 & ---\\
\bottomrule
\end{tabular}
\end{table*}

\section{Experiments}

\subsection{Datasets, Protocols, and Baselines}
\label{sec:datasets_protocols}

To ensure fair comparisons with state-of-the-art approaches, we follow the experimental setup of ~\cite{cermelli2020modelingthebackground} for datasets, protocol, metrics, and baseline implementations. ALthough, we also propose to evaluate on new datasets and on more challenging protocols. Furthermore, we explore in advanced experiments, for the first time, how rehearsal can improve performance in CSS.

\noindent\textbf{Datasets:\,} We evaluate our model on three datasets, summarized in \autoref{tab:dataset_description}: Pascal-VOC~\cite{everingham2015pascalvoc}, Cityscapes~\cite{cordts2016cityscapes} and ADE20k~\cite{zhou2017adedataset}. VOC contains 20 classes, 10,582 training images, and 1,449 testing images. Cityscapes contains 2975 and 500 images for train and test, respectively. Those images represent 19 classes and were taken from 21 different cities. ADE20k has 150 classes, 20,210 training images, and 2,000 testing images. All ablations and hyperparameters tuning were done on a validation subset of the training set made of 20\% of the images. 
For all datasets, we use random resize and crop augmentation (scale from 80\% to 110\%), as well as random horizontal flip during training time. The final image size for Pascal-VOC and ADE20k is $512 \times 512$ while it is $512 \times 1024$ for Cityscapes. 

\noindent\textbf{CSS protocols:\,} ~\cite{cermelli2020modelingthebackground} introduced the \textit{Overlapped} setting where only the current classes are labeled vs. a background class $\mcC^t$. Moreover, pixels can belong to any classes $\mcC^{1:t-1} \cup \mcC^{t} \cup \mcC^{t+1:T}$ (old, current, and future). Those two constraints make this setting both challenging and realistic, as in a real setting there isn't any oracle method to exclude future classes from the background. In all our experiments, we respect the Overlapped setting. While the training images are only labeled for the current classes, the testing images are labeled for all seen classes. We evaluate several CSS protocols for each dataset, e.g. on VOC 19-1, 15-5, and 15-1 respectively consists in learning 19 then 1 class ($T=2$ steps), 15 then 5 classes ($2$ steps), and 15 classes followed by five times 1 class ($6$ steps). The last setting is the most challenging due to its higher number of steps. Similarly, on Cityscapes 14-1 means 14 followed by five times 1 class ($6$ steps) and on ADE 100-50 means 100 followed by 50 classes ($2$ steps).
We provide a summary of all these settings in \autoref{tab:setting_description}.

\noindent\textbf{Metrics:\,} we compare the different models using traditional mean Intersection over Union (mIoU). Specifically, we compute mIoU after the last step $T$ for the initial classes $\mcC^{1}$, for the incremented classes $\mcC^{2:T}$, and for all classes $\mcC^{1:T}$ (\textit{all}). These metrics respectively reflect the robustness to catastrophic forgetting (the model rigidity), the capacity to learn new classes (plasticity), as well as its overall performance (trade-of between both). We also introduce a novel \textit{avg} metric (short for \textit{average}), which measures the average of mIoU scores measured step after step, integrating performance over the whole continual learning process. We stress that all four metrics are important and we shouldn't disregard one for the other: a model suffering no forgetting but not able to learn anything new is useless. 

\noindent\textbf{Baselines:\,} We benchmark our model against the latest state-of-the-arts CSS methods ILT~\cite{michieli2019ilt}, MiB~\cite{cermelli2020modelingthebackground}, GIFS \cite{cermelli2020fewshotcontinualsegm}, and SDR \cite{michieli2021sdr}. Note that while GIFS was created by Cermelli et al.~\cite{cermelli2020fewshotcontinualsegm} for continual \textit{few-shots} segmentation, we adapt it for the more general task of CSS. Unless stated otherwise, all the results are excerpted from the corresponding papers. We also evaluate general continual models based on weight constraints (PI \cite{zenke2017synaptic_intelligence}, EWC \cite{kirkpatrick2017ewc}, and RW \cite{chaudhry2018riemannien_walk}) and knowledge distillation (LwF \cite{li2018lwf} and LwF-MC \cite{rebuffi2017icarl}). Moreover, unless explicitely stated otherwise, all the models (ours included), do not use rehearsal learning.

\noindent\textbf{Implementation Details:\,}As in~\cite{cermelli2020modelingthebackground}, we use a Deeplab-V3~\cite{chen2017deeplabv3} architecture with a ResNet-101~\cite{he2016resnet} backbone pretrained on ImageNet~\cite{deng2009imagenet} for all experiments and models but SDR \cite{michieli2021sdr} which used a Deeplab-V3+~\cite{chen2018deeplabv3plus}. For all datasets, we set a maximum threshold for the uncertainty measure of Eq. 7 to $\tau=1e-3$. We train our model for 30, 60, and 30 epochs per CSS step on Pascal-VOC, ADE20k, and Cityscapes, respectively, with an initial learning rate of $1e-2$ for the first CSS step, and $1e-3$ for all the following ones. Note that for Cityscapes, the first step is longer with 50 epochs. We reduce the learning rate exponentially with a decay rate of $9e-1$. We use SGD optimizer with $9e-1$ Nesterov momentum. The Local POD weighting hyperparameter $\lambda$ is set to $1e-2$ and $5e-4$ for intermediate feature maps and logits, respectively. Moreover, we multiply this factor by the adaptive weighting $\sqrt{\nicefrac{|C^{1:t}|}{|C^{t}|}}$ introduced by \cite{hou2019ucir} that increases the strength of the distillation the further we are into the continual process. For all feature maps, Local POD is applied before ReLU, with squared pixel values, as in \cite{zagoruyko2016distillation_attention,douillard2020podnet}. We use 3 scales for Local POD: $1$, $\nicefrac{1}{2}$, and $\nicefrac{1}{4}$, as adding more scales experimentally brought diminishing returns. For PLOPLong only, we L2-normalize all POD embeddings before distilling them as also done on \cite{douillard2020podnet}. Furthermore for PLOPLong, the gradient norm-clipping is set at $1.0$ for Pascal-VOC and $2.0$ for Cityscapes. We use a batch size of 24 distributed on two 12Go Titan Xp GPUs. Contrary to many continual models, we don't have access to any task id in inference, therefore our setting/strategy has to predict a class among the set of all seen classes.

\subsection{Quantitative Evaluation}
\label{sec:quantitative}

\begin{figure}
  \includegraphics[width=0.9\linewidth]{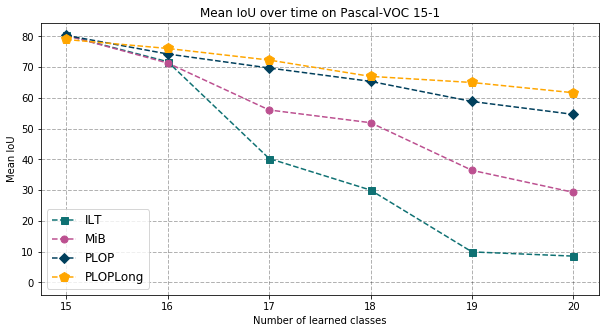} 
     \vspace*{-0.3cm}
    \caption{mIoU evolution on Pascal-VOC 2012 15-1. While MiB's mIoU quickly deteriorates, PLOP and PLOPLong's mIoU remains high, due to improved resilience to catastrophic forgetting.}
    \label{fig:plot_voc_15-1}
\end{figure}

\subsubsection{Pascal VOC 2012}
\autoref{tab:voc_sota1} shows quantitative experiments on VOC 19-1, 15-5, and 15-1. both the proposed PLOP and PLOPLong outperform state-of-the-art approaches, MiB \cite{cermelli2020modelingthebackground}, SDR \cite{michieli2021sdr}, and GIFS \cite{cermelli2020fewshotcontinualsegm} on all evaluated settings by a significant margin. For instance, on 19-1, the forgetting of old classes (1-19) is reduced by 4.39 percentage points (\pp) while performance on new classes is greatly improved (+13.76 \pp), as compared to the best performing method so far, MIB \cite{cermelli2020modelingthebackground}. On 15-5, our model is on par with our re-implementation of MiB, and surpasses the original paper scores \cite{cermelli2020modelingthebackground} by 1 \pp. On the most challenging 15-1 setting, general continual models (EWC and LwF-MC) and ILT all have very low mIoU. While MiB shows significant improvements, \ours still outperforms it by a wide margin. Furthermore, \ours also significantly outperform the best performing approach on this setting, GIFS \cite{cermelli2020fewshotcontinualsegm} (+6.11 \pp). Furthermore, mIoU for the joint model is $77.40\%$, thus \ours narrows the gap between CSS and joint learning on every CSS scenario. The average mIoU is also improved (+5.78 \pp) compared to GIFS, indicating that each CSS step benefits from the improvements related to our method. This is echoed by \autoref{fig:plot_voc_15-1}, which shows that while mIoU for both ILT and MiB deteriorates after only a handful of steps, \ours's mIoU remains very high throughout, indicating improved resilience to catastrophic forgetting and background shift. Last but not least, while PLOP performs better than PLOPLong on short setups (e.g. 19-1, 15-5), PLOPLong performs better, however, on longer CSS benchmarks such as 15-1 (+2.81 \pp). %We also include GIFS and SDR, two recent papers released after PLOP which also consider continual segmentation. While both significantly improve the mIoU of old classes, they remain largely inferior to PLOP and PLOPLong. Furthermore, we show an important gain in mIoU for new classes that is not present for neither SDR nor GIFS.

\begin{table}[t]
\centering
\caption{Continual Semantic Segmentation results on Cityscapes 14-1 in Mean IoU (\%).}
\vspace*{-0.3cm}
\label{tab:cityscapes_class}
\begin{tabular}{@{}l|p{1cm}p{1cm}|cccc@{}}
\toprule
 & \multicolumn{6}{c}{\textbf{14-1} (6 tasks)}\\
\textbf{Method} &  1-14 & 15-19 & \textit{all} & \textit{avg}\\
\midrule
%MiB OLD Version & --- & 0 & 60.34 & 14.42 & 48.86 & 55.01\\
%PLOP OLD Version & --- & 0 &  59.92 & 11.29 & 47.76 & 53.73\\
%PLOPv2 OLD Version & --- & 0 & 58.93 & 19.78 & 49.14 & 55.27\\
%PLOPv2 OLD Version & Image Rehearsal & 117.0 & \textbf{59.73} & 19.71 & 49.73 & 56.10 \\
%PLOPv2 OLD Version & Object Rehearsal & 0.8 & 59.12 & \textbf{37.05} & \textbf{53.60} & \textbf{57.25}\\
%\midrule
MiB \cite{cermelli2020modelingthebackground} & 55.11 & 12.91 & 44.56 & 49.76\\
\ours & 56.59 & 13.07 & 45.71 & 51.28\\
\ourslong & \textbf{58.60} & \textbf{15.04} & \textbf{47.71} & \textbf{54.31}\\
\bottomrule
\end{tabular}
\end{table}

\begin{table*}[t]
\centering
\caption{Continual Semantic Segmentation results on ADE20k in Mean IoU (\%).}
\vspace*{-0.3cm}
\label{tab:ade_sota}
\begin{tabular}{@{}l|cccc||cccc||cccc@{}}
\toprule
& \multicolumn{4}{c}{\textbf{100-50} (2 tasks)} & \multicolumn{4}{c}{\textbf{50-50} (3 tasks)} & \multicolumn{4}{c}{\textbf{100-10} (6 tasks)}\\
\textbf{Method} & 0-100 & 101-150 & \textit{all} & \textit{avg} & 0-50 & 51-150 & \textit{all} & \textit{avg}  & 0-100 & 101-150 & \textit{all} & \textit{avg} \\
\midrule
ILT \cite{michieli2019ilt} & 18.29  & 14.40  & 17.00  & 29.42  & \tableindent 3.53  & 12.85  & \tableindent 9.70 & 30.12 & \tableindent 0.11  & \tableindent 3.06 & \tableindent 1.09 & 12.56 \\ 
MiB \cite{cermelli2020modelingthebackground} & 40.52  & \textbf{17.17}  & \textbf{32.79}  & \textbf{37.31}  & 45.57  & \textbf{21.01}  & 29.31 & 38.98 & 38.21 & 11.12 & 29.24 & 35.12 \\
\ours  & \textbf{41.87}  & 14.89  & \textbf{32.94}  & \textbf{37.39}  & \textbf{48.83}  & \textbf{20.99}  & \textbf{30.40} & \textbf{39.42} & \textbf{40.48}  & \textbf{13.61} & \textbf{31.59} & \textbf{36.64}\\
% Note that ADE 50-50 was done without adaptive!
%\ourslong & \tbd  & \tbd  & \tbd  & \tbd  & \tbd  & \tbd & \tbd & \tbd  & 35.37 & 16.35 & 29.07 & 35.77 \\
%\midrule
%\ours vs MiB & +1.35  & -2.28  & +0.15  & +0.08  & +3.26  & -0.02  & +1.09 & +0.44 & +2.27  & +2.49 & +2.35 & +1.52 \\
%\midrule
% Joint model & 44.30 & 28.20 & 38.90 & --- & 51.10 & 33.25 & 38.90 & ---  & 44.30 & 28.20 & 38.90 & --- \\
\bottomrule
\end{tabular}
\end{table*}

\subsubsection{Cityscapes}
We also validate our method on Cityscapes in \autoref{tab:cityscapes_class}. In order to design a setting similar to Pascal-VOC 15-1 (6 tasks), we design the 14-1 setting. Also, we simulate a background class by folding together the unlabeled classes. Here again, PLOP performs slightly better than MIB (+2.48 \pp on the old classes, +0.16 \pp on the new ones, +1.15 \pp on average). Moreover, PLOPLong performs significantly better (+3.59 \pp on the old classes, +2.13 \pp on the new ones, +3.15 \pp on average), indicating better robustness to both catastrophic forgetting and background shift, especially when considering long continual learning setups. To more precisely assess this phenomenon, we investigate the performance of both methods when dealing with longer task learning sequences.

\subsubsection{ADE20k}
\autoref{tab:ade_sota} shows experiments on ADE 100-50, 100-10, and 50-50. This dataset is notoriously hard, as the joint model baseline mIoU is only 38.90\%. ILT has poor performance in all three scenarios. \ours shows comparable performance with MiB on the short setting 100-50 (only 2 tasks), improves by 1.09 \textit{p.p} on the medium setting 50-50 (3 tasks), and significantly outperforms MiB with a wider margin of 2.35 \textit{p.p} on the long setting 100-10 (6 tasks). In addition to being better on all settings, PLOP showcased an increased performance gain on longer CSS (e.g. 100-10) scenarios, due to increased robustness to catastrophic forgetting and background shift.

\begin{table}[t]
\centering
\caption{Mean IoU on Pascal-VOC 2012 10-1.}
\vspace*{-0.3cm}
\label{tab:voc_hard}
\begin{tabular}{@{}l|cccc@{}}
\toprule
 & \multicolumn{4}{c}{\textbf{VOC 10-1} (11 tasks)}\\
\textbf{Method} & 0-10 & 11-20 & \textit{all} & \textit{avg}\\
\midrule
ILT \cite{michieli2019ilt} & \tableindent 7.15 & \tableindent 3.67 & \tableindent 5.50 & 25.71\\ 
MiB \cite{cermelli2020modelingthebackground} & 12.25 & 13.09 & 12.65 & 42.67 \\
\ours & 44.03 & 15.51 & 30.45 & 52.32\\
\ourslong & \textbf{61.06} & \textbf{18.56} & \textbf{40.83} & \textbf{58.62}\\
%{\color{red}\ourslong + OR} & \textbf{59.82} & \textbf{22.28} & \textbf{41.95} & \textbf{58.74}\\
\bottomrule
\end{tabular}
\end{table}

\begin{table}[t]
\centering
\caption{Mean IoU on ADE20k 100-5.}
\vspace*{-0.3cm}
\label{tab:ade_hard}
\begin{tabular}{@{}l|cccc@{}}
\toprule
 & \multicolumn{4}{c}{\textbf{ADE 100-5} (11 tasks)}\\
\textbf{Method} & 0-100 & 101-150 & \textit{all} & \textit{avg}\\
\midrule
ILT \cite{michieli2019ilt} & \tableindent 0.08 & \tableindent 1.31 & \tableindent 0.49 & \tableindent 7.83\\
MiB \cite{cermelli2020modelingthebackground} & 36.01 & \tableindent 5.66 & 25.96  & 32.69 \\
\ours  & \textbf{39.11} & \tableindent \textbf{7.81} & \textbf{28.75} & \textbf{35.25}\\
% \ourslong & 24.78 & 10.23 & 19.97 & 30.03\\
\bottomrule
\end{tabular}
\end{table}

\subsubsection{Longer Continual Learning}
We argue that CSS experiments should push towards more steps \cite{wortsman2020supermasks,lomonaco2020ar1,douillard2020podnet,castro2018end_to_end_inc_learn} to quantify the robustness of approaches w.r.t. catastrophic forgetting and background shift. We introduce two novel and much more challenging settings with 11 tasks, almost twice as many as the previous longest setting. We report results for VOC 10-1 in \autoref{tab:voc_hard} (10 classes followed by 10 times 1 class) and ADE 100-5 in \autoref{tab:ade_hard} (100 classes followed by 10 times 5 classes). The second previous State-of-the-Art method, ILT, has a very low mIoU ($<6$ on VOC 10-1 and practically null on ADE 100-5). Furthermore, the gap between \ours and MiB is even wider compared with previous benchmarks (e.g. $\times$3.6 mIoU on VOC for mIoU of base classes 1-10), which confirms the superiority of \ours when dealing with long continual processes. Furthermore, in such a case, PLOPlong really shines, bringing significant improvements (+17.03 \pp on the old classes, +3.05 \pp on the new classes, +10.42 \pp on average) due to combination of cosine normalization (both on the classifier and Local POD) and its frozen BatchReNormalization.

\subsection{Object Rehearsal}

\begin{table*}[t]
\centering
\caption{Comparison of rehearsal-based methods on Pascal-VOC 2012 15-1 overlap in Mean IoU (\%). We only consider the time overhead spent after the first task whose computation overhead is similar for all methods.}
\vspace*{-0.3cm}
\label{tab:voc_rehearsal_learning}
\begin{tabular}{@{}l|ccc|cccc@{}}
\toprule
 & \multicolumn{7}{c}{\textbf{15-1} (6 tasks)}\\
\textbf{Method} & \textbf{Rehearsal} & \textbf{Memory} (Mb) $\downarrow$ & \textbf{Time} (Hours) $\downarrow$ &  0-15 & 16-20 & \textit{all} & \textit{avg}\\
\midrule
\ours  & --- & 0 & 1.8 & 65.12 & 21.11 & 54.64 & 67.21\\
\ourslong & --- & 0 & 1.8 & 72.00 & 26.66 & 61.20 & 70.02\\
%HRHF \cite{huang2021halfrealhalffake} & DeepInversion & 0 & 5.9 & 72.40 & 39.60 & 64.60 & \\
\hdashline
Yu et al.\cite{yu2020continualsegmentationselftraining} & Unlabeled COCO & 20,000 & 7.0 & 71.40 & 40.00 & 63.60 &  \\
\ours & Unlabeled COCO & 20,000 & 1.4 & 72.57 & 45.08 & 66.03 & 71.85 \\
\ours & Unlabeled VOC & 2,000 & 1.4 &  75.32 & 52.59 & 69.91 & 75.21\\
%PLOP & Partial VOC & 2,000 & \tbd & 74.91 & 54.55 & 70.06 & 75.41 \\
%PLOPv2 & Partial VOC & 2,000  & \tbd & 76.87 & 56.07 & 71.92 & 74.88\\
\hdashline
\ourslong & Partial VOC & 2.2 & 2.6 & 74.14  & 38.87 & 65.74 & 72.02\\
\ourslong & Partial VOC & 22 & 2.6 & \textbf{74.18} & 43.22 & 66.81 & \textbf{72.48}\\
\ourslong & Object VOC & 0.26 & 2.7 & 73.32 & 42.86 & 66.07 & 72.21\\
\ourslong & Object VOC & 2.6 & 2.7 & 73.79 & \textbf{45.78} & \textbf{67.12} & \textbf{72.42}\\
\midrule
Joint model & --- & --- & --- & 79.10 & 72.60 & 77.40 & ---\\
\bottomrule
\end{tabular}
\end{table*}

\noindent\textbf{Pascal-VOC 15-1:\,} We now allow CSS models to store information from the previous steps and classes: in such a case, the overall model efficiency can be understood as to what extent it allows to find good trade-off between its accuracy (as measured by the aforementionned standard metrics) and the memory footprint of the stored images or objects. Methods such as HRHF \cite{huang2021halfrealhalffake} can not easily be understood in these terms as they do not store data, strictly speaking, but rather use deep inversion \cite{yin20deepinversion} techniques to generate synthetic data, which comes at a high time requirement: thus, we exclude this method from our comparisons. We first consider the challenging Pascal-VOC 15-1 setting in \autoref{tab:voc_rehearsal_learning}. We use PLOP and PLOPLong as baselines with 0 memory overhead, as both models only use data from the current task. Moreover, we compare with \cite{yu2020continualsegmentationselftraining}, where an unlabeled external dataset such as COCO \cite{lin2014mscocodataset} is used through pseudo-labelling to improve performance on Pascal-VOC, as both datasets present significant overlap in terms of classes and domains. We reimplemented their method and also considered PLOP in this configuration. Using the external COCO provide PLOP an important gain of mIoU for both old classes (+7 \textit{p.p.}) and new classes (+24 \textit{p.p.}). Furthermore, PLOP with COCO is significantly more performant than Yu et al. model (+5.43 \textit{p.p.}) despite the latter was designed explicitly to use such unlabeled external dataset. Notice also how PLOPLong, without any kind of rehearsal, remains equivalent to \cite{yu2020continualsegmentationselftraining} in terms of mIoU on 0-15 (72.00\% vs 71.40\%), despite the latter using a large pool of data (+20Go). Perhaps counter-intuitively the gain is located on the new classes (26.66\% vs 40.00\%) as the rehearsal effect has a regularizing effect leading to less over-predicting of the recent classes. A drawback of using COCO is that the visual domain is not exactly the same as VOC, therefore we also considered using PLOP with a rehearsal of the unlabeled VOC. Without surprise, it results in a much better overall performance (+3.88 \textit{p.p.} compared to using COCO).
Another setup that we consider is the image rehearsal paradigm, where, at each step, we keep a number of (randomly selected) images along with their segmentation map. These segmentation maps are however incomplete, due to the nature of the CSS problem (see \autoref{sec:object_rehearsal}): hence, we refer to this setting as partial VOC. We consider two amounts of images to keep: 10 images per class (22 Mb) and 1 image per class (2.2 Mb). PLOPLong largely benefits from this rehearsal, most notably on new classes (16-20) with a gain up to 16.56 \textit{p.p.}. Finally, we compare our novel Object Rehearsal denoted by ``\textit{object VOC}'' where we store either one object per class (0.26 Mb) or 10 objects per class (2.6 Mb). As shown on \autoref{tab:voc_rehearsal_learning}, The proposed object rehearsal, in addition to being significantly more memory efficient than whole image rehearsal (8.5$\times$ less space used), is equivalent or better in terms of mIoU especially for new classes (2.56 \textit{p.p.}). The ensemble of our experiments proves that rehearsal, when the partial or missing labeling is carefully handled, can provide important performance gain. Furthermore, our novel object rehearsal manages to strike the best trade-off with a minimal memory overhead without impacting its performance. 

\begin{comment}
\begin{figure}
    \centering
    \includegraphics[width=0.7\linewidth]{images/object_pasting.pdf}
    %\vspace*{-0.3cm}
    \caption{Pasted object for rehearsal in three images. Top row shows the pasting of a \texttt{motorbike} and a \texttt{bicycle} in Pascal-VOC, and bottom row shows the pasting of a \texttt{bus} in Cityscapes. \textcolor{green}{A garder?}}
    \label{fig:object_pasting}
\end{figure}
\end{comment}

\noindent\textbf{Cityscapes:\,} We propose more experiments with Object Rehearsal in \autoref{tab:cityscapes_rehearsal} where we apply our model on Cityscapes 14-1.
For the rehearsal, we sample either 10 images or object per class. Cityscapes images are particularly large (even resized to $512 \times 1024$)  and "empty" (most of it is road and sky). Consequently the memory overhead is extremely important compared object rehearsal (117 vs 0.8). We show that both our novel image and object rehearsal improve performance of the already competitive PLOPLong, and with object rehearsal providing the large gain (+9 \textit{p.p.} in \textit{all}).

\begin{table}[t]
\centering
\caption{Continual Semantic Segmentation results on Cityscapes 14-1 overlap in Mean IoU (\%).}
\vspace*{-0.3cm}
\label{tab:cityscapes_rehearsal}
\begin{tabular}{@{}l|p{1cm}p{1cm}|cccc@{}}
\toprule
 & \multicolumn{6}{c}{\textbf{14-1} (6 tasks)}\\
\textbf{Method} & \textbf{Rehearsal} & \textbf{Memory} &  1-14 & 15-19 & \textit{all} & \textit{avg}\\
\midrule
%MiB OLD Version & --- & 0 & 60.34 & 14.42 & 48.86 & 55.01\\
%PLOP OLD Version & --- & 0 &  59.92 & 11.29 & 47.76 & 53.73\\
%PLOPv2 OLD Version & --- & 0 & 58.93 & 19.78 & 49.14 & 55.27\\
%PLOPv2 OLD Version & Image Rehearsal & 117.0 & \textbf{59.73} & 19.71 & 49.73 & 56.10 \\
%PLOPv2 OLD Version & Object Rehearsal & 0.8 & 59.12 & \textbf{37.05} & \textbf{53.60} & \textbf{57.25}\\
%\midrule
%MiB \cite{cermelli2020modelingthebackground} & --- & 0 & 55.11 & 12.91 & 44.56 & 49.76\\
%\ours & --- & 0 &  56.59 & 13.07 & 45.71 & 51.28\\
\ourslong & --- & 0 & 58.60 & 15.04 & 47.71 & 54.31\\
\ourslong & Partial & 117.0 & \textbf{58.93} & 19.55 & 49.09 & \textbf{55.74} \\
\ourslong & Object & 0.8 & 57.82 & \textbf{23.13} & \textbf{49.15} & 54.80\\
\bottomrule
\end{tabular}
\end{table}

\subsection{Model Introspection}

In this section, we perform component by component ablation study of PLOP, but also aim at understanding the behavior of the model, as well as comparing different rehearsal alternatives. Lastly we propose qualitative visualizations of the models predictions.

\subsubsection{PLOP Ablations}

We compare several distillation and classification losses on VOC 15-1 to stress the importance of the components of \ours and report results in \autoref{tab:ablation_distill_classif}. All comparisons are evaluated on a val set made with 20\% of the train set, therefore results are slightly different from the main experiments.

\noindent\textbf{Distillation comparisons:\,}\autoref{tab:ablation_distillation} compares different distillation losses when combined with our pseudo-labeling loss. As such, UNKD introduced in \cite{cermelli2020modelingthebackground} performs better than the Knowledge Distillation (KD) of \cite{hinton2015knowledge_distillation}, but not at every step (as indicated by the \textit{avg.} value), which indicates instability during the training process. POD, proposed in \cite{douillard2020podnet}, improves the results on the old classes, but not on the new classes (16-20). In fact, due to too much plasticity, POD model likely overfits and predicts nothing but the new classes, hence a lower mIoU.  Finally, Local POD leads to superior performance (+20 \pp) w.r.t. all metrics, due to its integration of both long and short-range dependencies. This final row represents our full \ours strategy.

\noindent\textbf{Classification comparisons:\,}\autoref{tab:ablation_classif} compares different classification losses when combined with our Local POD distillation loss. Cross-Entropy (CE) variants perform poorly, especially on new classes. UNCE, introduced in \cite{cermelli2020modelingthebackground}, improves by merging the background with old classes, however, it still struggles to correctly model the new classes, whereas our pseudo-labeling propagates more finely information of the old classes, while learning to predict the new ones, dramatically enhancing the performance in both cases. This penultimate row represents our full \ours strategy.
Also notice that the performance for pseudo-labeling is very close to \textit{Pseudo-Oracle} (where the incorrect pseudo-labels are removed), which may constitute a performance ceiling of our uncertainty measure. A comparison between these two results illustrates the relevance of our entropy-based uncertainty estimate.

\subsubsection{Robutness to class ordering}

\begin{figure}
    \includegraphics[width=0.9\linewidth]{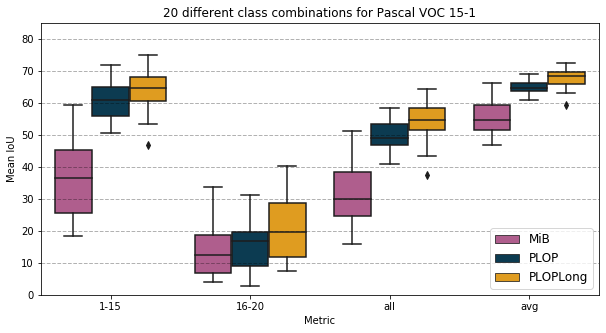}
    \vspace*{-0.3cm}
    \caption{Boxplots of the mIoU of initial classes (1-15), new (16-20), all, and average for 20 random class orderings. PLOP is significantly better and more stable than MiB. PLOPLong further improves upon PLOP by better retaining old class information.}
    \label{fig:order_voc_15-1}
\end{figure}

Continual learning methods may be prone to instability. It has already been shown in related contexts \cite{kim2019medic} that class ordering can have a large impact on performance. Unfortunatly, in real-world settings, the optimal class order can never be known beforehand: thus, the performance of an ideal CSS method should be as invariant to class order as possible. In all experiments done so far, this class order has been kept constant, as defined in~\cite{cermelli2020modelingthebackground}. We report results in \autoref{fig:order_voc_15-1} as boxplots obtained by applying 20 random permutations of the class order on VOC 15-1. We report in \autoref{fig:order_voc_15-1} (from left to right) the mIoU for the old, new classes, all classes, and average over CSS steps. In all cases, \ours surpasses MiB in term of avg mIoU. Furthermore, the standard deviation (e.g. 10\% vs 5\% on \textit{all}) is always significantly lower, showing the excellent stability of \ours compared with existing approaches. Furthermore, PLOPLong, while having more variance on the new classes, performs better overall as compared to PLOP, not to mention MiB. This is due to the fact that PLOPLong, due to improved distillation loss (at prediction and Local POD levels) and batch re-normalization that allows to better retain information of the old classes, which again is more conspicuous on longer CSS scenarios.

\begin{table}
\centering
\caption{Comparison studies on Pascal-VOC 2012 15-1 on a validation subset of 20\% of the training set.}
%\vspace*{-0.3cm}
\label{tab:ablation_distill_classif}
\begin{subtable}{0.5\textwidth}
    \centering
    \caption{Pseudo loss (\autoref{eq:pseudo_loss}) with different distillation losses.}
    %\vspace*{-0.2cm}
    \label{tab:ablation_distillation}
    \begin{tabular}{@{}l|cccc@{}}
    \toprule
    Distillation loss & 0-15 & 16-20 & \textit{all} & \textit{avg}\\
    \midrule
    %ILT \cite{michieli2019ilt}'s distill & 19.91 & \tableindent 5.49 & 16.48 & 49.43\\
    %CSC \cite{park2020csc} & 25.49 & \tableindent 4.72 & 20.48 & 44.97\\
    Knowledge Distillation & 29.72 & \tableindent 4.42 & 23.69 & 49.18\\
    UNKD & 34.85 & \tableindent 5.26 & 27.80 & 46.39\\
    POD & 43.94 & \tableindent 4.82 & 34.62 & 53.35\\
    Local POD (\autoref{eq:local_pod_loss}) & \textbf{63.06} & \textbf{17.92} & \textbf{52.31} & \textbf{65.71}\\
    \bottomrule
    \end{tabular}
\end{subtable}
\hfill
\vspace{0.5cm}
\begin{subtable}{0.5\textwidth}
    \centering
    \caption{Local POD loss (\autoref{eq:local_pod_loss}) with different classification losses.}
    %\vspace*{-0.2cm}
    \label{tab:ablation_classif}
    \begin{tabular}{@{}l|cccc@{}}
    \toprule
    Classification loss & 0-15 & 16-20 & \textit{all} & \textit{avg}\\
    \midrule
    CE only on new & 12.95 & \tableindent 2.54 & 10.47 & 47.02 \\
    CE & 33.80 & \tableindent 4.67 & 26.87 & 50.79 \\
    UNCE & 48.46 & \tableindent 4.82 & 38.62 & 53.19 \\
    Pseudo (\autoref{eq:pseudo_loss}) & \textbf{63.06} & \textbf{17.92} & \textbf{52.31} & \textbf{65.71}\\
    \midrule
    \textit{\small{Pseudo-Oracle}} & \textit{63.69} & \textit{23.35} & \textit{54.09} & \textit{66.05}\\
    %\textit{\small{Pseudo + corrected}} & \textit{66.88} & \textit{16.88} & \textit{54.98} & \textit{66.50}\\
    %\textit{\small{CE + all labels}}  & \textit{71.45} & \textit{10.78} & \textit{57.00} & \textit{67.04}\\
    \bottomrule
    \end{tabular}
\end{subtable}
\end{table}

\subsubsection{Rehearsal alternatives}

\autoref{tab:rehearsal_alternative} draws a comparison between several rehearsal alternatives in CSS. We split methods according to three criterions: ``\textit{type}'' denotes whether we store a whole image, an object, or a patch For both object and patch, we only rehearse a small amount of the images: in object, only the pixel's objects are used while for patch, we use the pixel's object and the close surrounding pixels that contain background information. We insert the reheasal data according to ``\textit{mixing}'', either by pasting (see \autoref{sec:object_rehearsal}), by mixing pixels according to mixup \cite{hingyi2018mixup} rule, or in the case of image rehearsal no mixing is done. Finally we consider two erasing methods: either all pixels are erased (\textit{All}), or only pixels belonging to non-background classes are erased (\textit{Foreground}). Note that, for the latter method, it includes classes detected via pseudo-labeling. For all methods, we randomly select 10 images/patches/objects per class for rehearsal. Without surprise, Patch (III-V) and Object-based (VI-X) rehearsal are more data-efficient than images (I and II) (4.50 and 2.60 vs 22.20 Mb). Mixing the rehearsed data with mixup (VI and VII) is less efficient than direct pasting (III, IV, and VIII-X) because segmentation, contrary to classification, requires sharp boundaries \cite{chen2020semeda}. We considered rehearsing each patch/object without mixing, but because their shape may variate no batching is possible which would be much slower. Therefore, we investigate mixing patches/objects while erasing all others pixels (III and VIII). This doesn't work, which is probably linked to the altered statistics of the batch normalization. On the other hand, we found that a selective erasing that remove foreground objects (V, VII, and X) while keeping the background of the destination image proved to be very effective (63.12\% to 67.12\% \textit{all} mIoU for object). This confirmed our intuition that pasting in segmentation is a delicate operation that may lead to confusion in the network, particularly on object boundaries. The erased pixels are replaced by gray pixels. We considered more complicated approaches such as in-painting \cite{fang2019instaboost} or textures filling \cite{mallikarjuna2006kth-tips}, but were slightly less effective than our simpler solution. Overall, through careful design, we manage to get the best performance using our novel object rehearsal which was also the most memory-efficient.

\begin{table}[t]
\centering
\caption{Rehearsal alternative on Pascal-VOC 2012 in Mean IoU (\%). Object/Patch-based methods with 10 objects/patches per class, and Image-based with 10 images per class. All experiments done with \ourslong.}
\vspace*{-0.3cm}
\label{tab:rehearsal_alternative}
\begin{tabular}{@{}llc|c|cc|r@{}}
\toprule
 & & & & \multicolumn{2}{c}{\textbf{15-1} (6 tasks)}\\
\textbf{Type} & \textbf{Mixing} & \textbf{Erase} & \textbf{Memory} $\downarrow$ & \textit{all} & \textit{avg} & \\
\midrule
\multirow{2}{*}{Image} & Mixup & ---         & \multirow{2}{*}{22.20} & 61.77 & 69.88 & I\\
                       & \,\,\,\,\,--- & --- &  & 66.81 & \textbf{72.48} & II\\
\hline
\multirow{3}{*}{Patch} & Pasting & All & \multirow{3}{*}{4.50} & 55.45 & 66.35 & III\\
                       & Pasting & --- &  & 63.41 & 70.75 & IV\\
                       & Pasting & Foreground &  & 66.28 & 71.66 & V\\
\hline
\multirow{5}{*}{Object} & Mixup & --- & \multirow{5}{*}{\textbf{2.60}} & 63.25 & 70.91 & VI\\
                        & Mixup & Foreground &  & 64.45 & 71.65 & VII\\
                        & Pasting & All &  &  52.26 & 65.97 & VIII\\
                        & Pasting & --- &  & 63.12 & 70.52 & IX\\
                        & Pasting & Foreground & &  \textbf{67.12} & \textbf{72.42} & X\\
\bottomrule
\end{tabular}
\end{table}

\begin{figure}
  \centering
  \includegraphics[width=\linewidth]{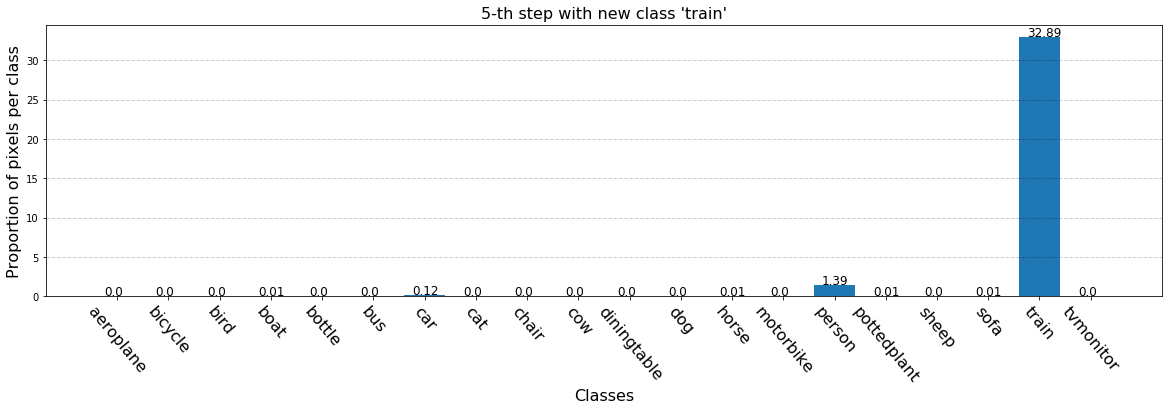} 
     \vspace*{-0.3cm}
    \caption{Distribution of the pixel per class during the $5^{\text{th}}$ step of Pascal-VOC 15-1.}
    \label{fig:distribution_voc_5th}
\end{figure}

\begin{figure*}
  \centering
  \includegraphics[width=\linewidth]{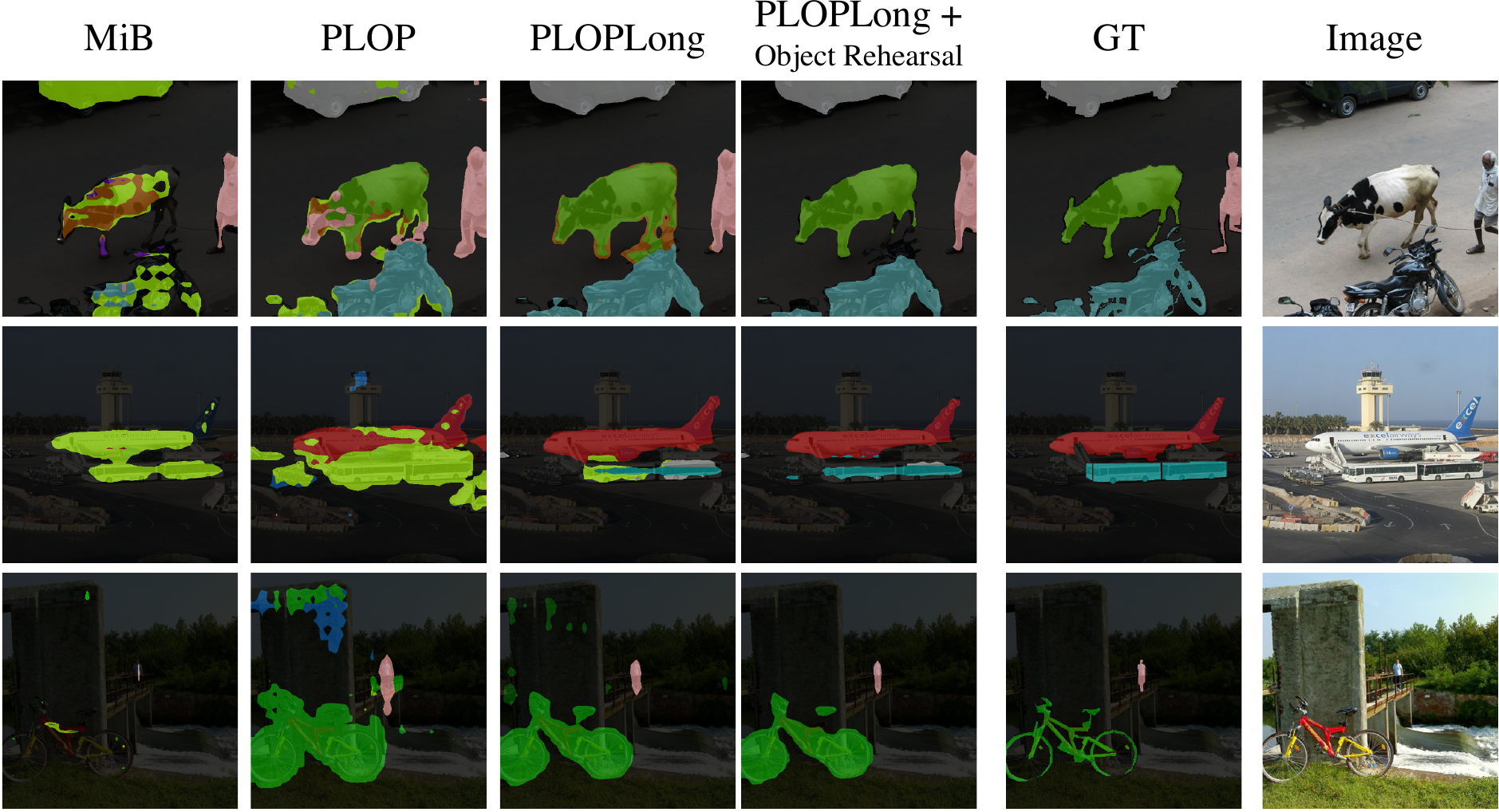}
    \caption{Visualization of the predictions of MiB, PLOP, PLOPLong, and PLOPLong with Object Rehearsal on three test images at the $6^\text{th}$ and final step on VOC 15-1 scenario. The first image contains \texttt{car}, \texttt{cow}, and \texttt{person}; the second \texttt{plane} and \texttt{bus}, and the third \texttt{bicycle} and \texttt{person}. While MiB doesn't manage to predict the correct classes, and tends to overpredict the most recent ones (e.g. \texttt{train} in light green). PLOP mostly grasp the correct classes, though sometimes with imprecision. PLOPLong and \textit{a fortiori} PLOPLong + Object rehearsal captures all existing classes, with the latter predicting almost perfect segmentation masks, compared to the ground-truth.}
    \label{fig:visualization}
\end{figure*}

\begin{figure}
    \centering
  \includegraphics[width=0.8\linewidth]{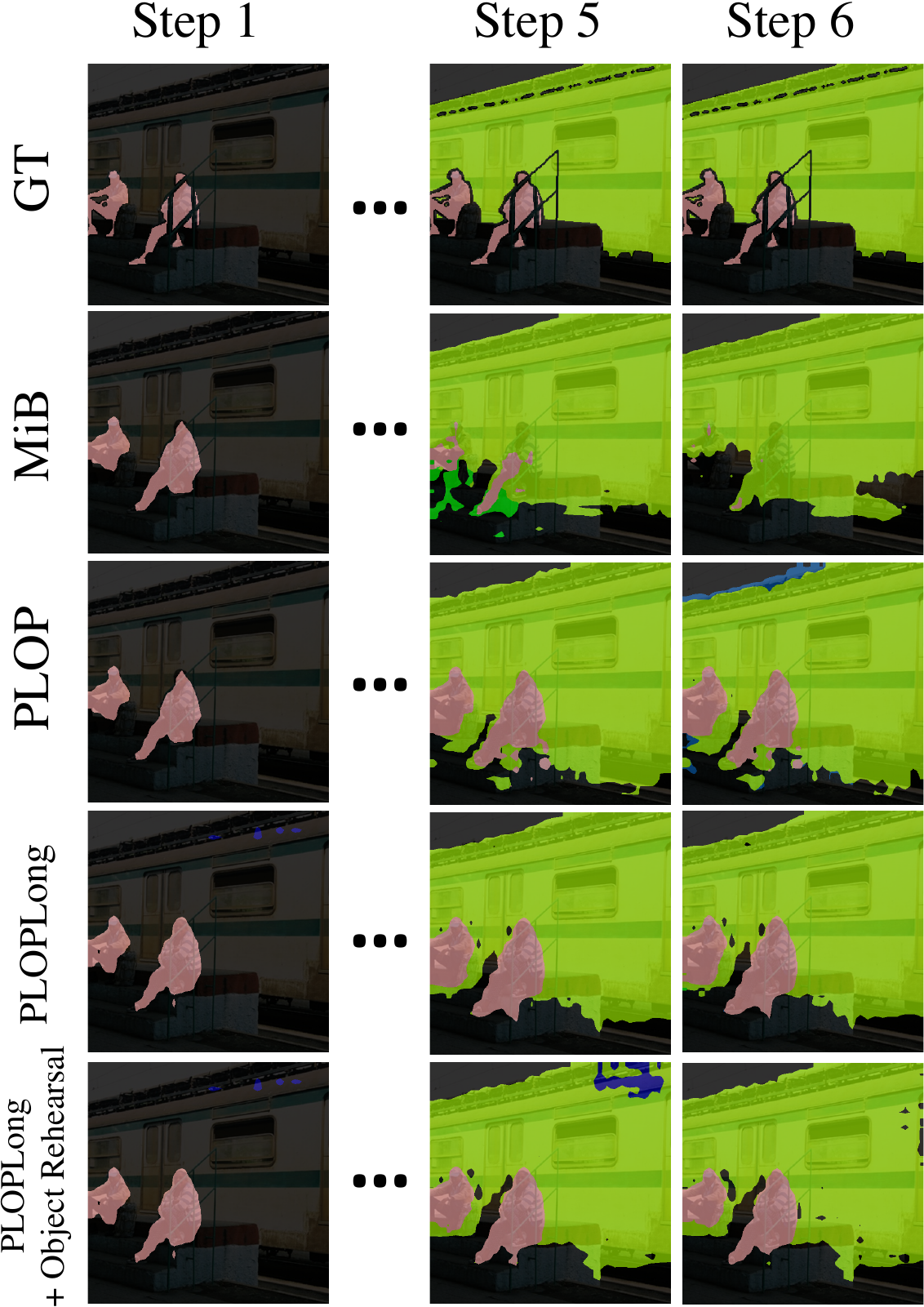}
    \caption{Visualization of the predictions of MiB, PLOP, PLOPLong, and PLOPLong with Object Rehearsal across time in VOC 15-1 on a test set image. At steps 1-4 only class \texttt{person} has been seen. At step 5, the class \texttt{train} is introduced, causing dramatic background shift. While MiB overfits on the new class and forget the old class, PLOP is able to predict both classes correctly. PLOPLong and PLOPLong + Object Rehearsal futher refine the predicted masks, resulting in much sharper boundaries.}
    \label{fig:visualization_gt_shift}
\end{figure}

\noindent\textbf{Rehearsal allievates pseudo-labeling's limitation:\,} All previous methods, PLOP and PLOPLong included, didn't consider rehearsal-learning. While popular in image classification, it has fewly been explored for continual segmentation. An attentive reader may wonder if rehearsal learning can have benefit given the fact that our pseudo-labeling can uncover ``hidden'' old classes in the images. Even a perfect pseudo-labeling (considered in \autoref{tab:ablation_classif}) may fail in some particular data situation where the class distribution of a step is almost dirac. We show in \autoref{fig:distribution_voc_5th} the distribution of pixels per class at the $5^{\text{th}}$ step of Pascal-VOC 15-1. The new class to learn, \texttt{train}, is abundant but also almost alone but the class \texttt{person}. In this case, no amount of pseudo-labeling can uncover previous classes like \texttt{bicycle} or \texttt{cat}. Our proposed rehearsal learning is complementary to our pseudo-labeling loss where the former can reduce the weakness for the latter in some cases.

\subsubsection{Visualization}

\autoref{fig:visualization} shows the predictions at the $6^\text{th}$ and final step of VOC 15-1 for four different models: MiB, PLOP, PLOPLong, and PLOPLong with Object Rehearsal. MiB struggles to even find the correct classes: in the first image the \texttt{car} is predicted as \texttt{train}, in the second the \texttt{plane} as \texttt{train}, and in the third no classes are detected. On the other hand, PLOP always manages to pick up the present classes, although sometimes with imperfection. PLOPLong further refines results, and the addition of Object Rehearsal produces a sharp and near perfect masks compared to the ground-truths (GTs). We showcased the harmful effect of background shift in \autoref{fig:visualization_gt_shift} where the class \texttt{person} is learned at $1^\text{st}$ step and the class \texttt{train} at $5^\text{th}$ step. MiB completly forgets the former class and overpredict the latter. The background shift is efficiently mitigated with pseudo-labeling as showed by the third row of PLOP. Our efficient proposition of object rehearsal allows further refinement of the predicted masks resulting in sharper boundaries.
\section{Conclusion}

Continual Semantic Segmentation (CSS) is an emerging but challenging computer vision domain. In this paper, we highlited two major issues in CSS: catastrophic forgetting and background shift. To deal with the former, we proposed Local POD, a multi-scale distillation scheme that preserves both long and short-range spatial statistics between pixels. This lead to to an effective balance between rigidity and plasticity for CSS, which in turns alleviate catastrophic forgetting. We then tackled background shift with an efficient uncertainty-based pseudo-labeling loss. It completes the partially-labeled segmentation maps, allowing the network to efficiently retain previously learned knowledge. Afterwards, We showed that carefully designed structural changes to the model could improve performance on long CSS scenarios, namely a cosine normalization adaptation of the classifier and Local POD followed by a modified batch normalization. Finally, we proposed to introduce rehearsal learning to CSS, one based on partially-labeled whole image rehearsal, and the other --much more memory-efficient-- consisting in object rehearsal. The latter further refining our already effective model performance and allowing real world application of CSS with a stringent memory constraint. We evaluated the proposed PLOPLong, with or without object rehearsal, on three datasets and over twelve different benchmarks. In each, we showed that our model performs significantly better than all existing baselines. Finally, we qualitatively validate our model through extensive ablations in order to better understand the performance gain. 

% or


\ifCLASSOPTIONcompsoc
  \section*{Acknowledgments}
\else
  \section*{Acknowledgment}
\fi

This  work  was  granted  access  to  the  HPC  resources  of IDRIS under the allocation AD011011706 made by GENCI. We thank Fabio Cermelli for the insightful discussions and his help with the BatchRenormalization code. We also thank Estelle Thou and Thomas Robert for the invaluable discussions and support.

\ifCLASSOPTIONcaptionsoff
  \newpage
\fi



%\bibliographystyle{IEEEtran}
\bibliographystyle{IEEEtran}
\bibliography{arthurcitations}

\begin{IEEEbiography}[{\includegraphics[width=1in,height=1.25in,clip,keepaspectratio]{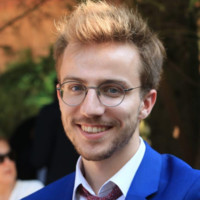}}]{Arthur Douillard}
is a second year PhD student at Sorbonne University in the MLIA / LIP6 team under the supervision of Prof. Matthieu Cord. He is also a research scientist at the fashion tech startup Heuritech, and a lecturer at the engineering school EPITA. He received a M.Eng. degree in Computer Science from EPITA in 2018. His research interests include continual learning, metric learning, segmentation, and more broadly computer vision.
\end{IEEEbiography}

\begin{IEEEbiography}[{\includegraphics[width=1in,height=1.25in,clip,keepaspectratio]{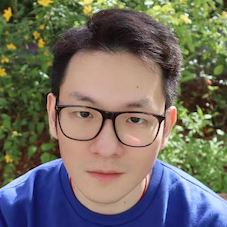}}]{Yifu Chen}
received his Ph.D. in 2020 from the Computer Science Laboratory (LIP6) of Sorbonne University, under the supervision of Prof. Matthieu Cord. His thesis is on deep learning methods for semantic image segmentation. His research also focuses on the visualization and explanation of deep neural networks, as well as on continual learning.
\end{IEEEbiography}

\begin{IEEEbiography}[{\includegraphics[width=1in,height=1.25in,clip,keepaspectratio]{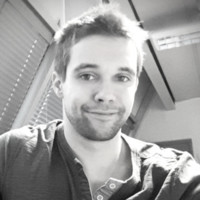}}]{Arnaud Dapogny} is a computer vision researcher at Datakalab in Paris. He obtained the Engineering degree from the Sup\'elec engineering School in 2011 and the Masters degree from Sorbonne University, Paris, in 2013 with high honors. He also obtained his PhD at Institute for Intelligent Systems and Robotics (ISIR) in 2016 and worked as a post-doctoral fellow at LIP6. His works concern deep learning for computer vision and its application to automatic facial behavior as well as gesture analysis.
\end{IEEEbiography}

\begin{IEEEbiography}[{\includegraphics[width=1in,height=1.25in,clip,keepaspectratio]{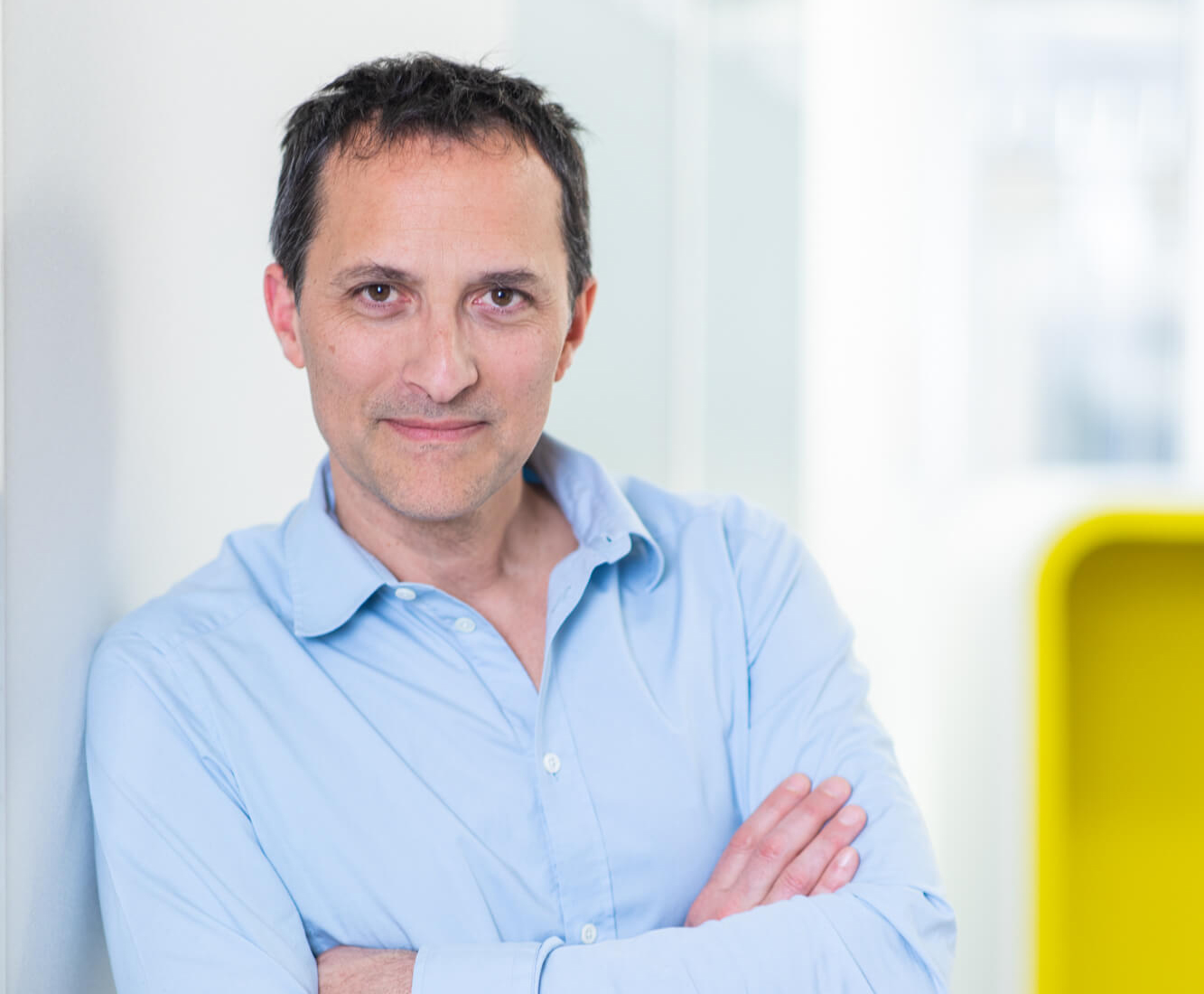}}]{Matthieu Cord}
is full professor at Sorbonne University. He is also part-time principal scientist at Valeo.ai. His research expertise includes computer vision, machine learning and artificial intelligence. He is the author of more 150 pub- lications on image classification, segmentation, deep learning, and multimodal vision and lan- guage understanding. He is an honorary member of the Institut Universitaire de France and served from 2015 to 2018 as an AI expert at CNRS and ANR (National Research Agency).
\end{IEEEbiography}

%\appendices
%\input{content/supplementary}

\end{document}